%% file: lf-draft-elsarticle.tex
\documentclass[preprint]{elsarticle}
\RequirePackage[OT1]{fontenc}
\RequirePackage{amsthm,amsmath}
\RequirePackage{natbib}
\RequirePackage[colorlinks,citecolor=blue,urlcolor=blue]{hyperref}

\usepackage[margin=0.9in]{geometry}
\usepackage{mathrsfs}
\usepackage{amssymb}
\usepackage{epsfig}
\usepackage{url}
\usepackage{amsmath}
\usepackage{natbib}
\usepackage{amsthm}
\usepackage{booktabs}
\usepackage{url}
\usepackage{alltt}
\usepackage{enumerate}
\usepackage{multirow}
\usepackage{bm}
\usepackage{color}
\usepackage{algpseudocode}
\usepackage{algorithm}

\newcommand{\vx}{\mathbf{x}}
\newcommand{\vy}{\mathbf{y}}

\newcommand{\Real}{I \! \! R}

\newcommand{\sD}{\mathscr{D}}

\newcommand{\vw}{\mathbf{w}}

\input Macros

\newtheorem{theorem0}{Theorem}
\newtheorem{lemma0}{Lemma}
\newtheorem{remark0}{Remark}
\newtheorem{fact0}{Fact}
\newtheorem{example0}{Example}
\newtheorem{definition0}{Definition}
\newtheorem{corollary0}{Corollary}
\newtheorem{proposition0}{Proposition}
\newtheorem{algorithmY}{Algorithm}
\newtheorem{conjecture0}{Conjecture}
\newtheorem{sketch0}{Sketch}

\newenvironment{conjecture}{\begin{conjecture0} \mbox{} }{\end{conjecture0}}
\newenvironment{sketch}{\begin{sketch0} \mbox{} }{\end{sketch0}}

\usepackage{lineno,hyperref}
\modulolinenumbers[5]

\journal{}

\begin{document}

\begin{frontmatter}

\title{Random Subspace Learning Approach to High-Dimensional Outliers Detection}

\author{Bohan Liu\fnref{myfootnote}}
\author{Ernest Fokou\'e\fnref{myfootnote}}

\address{Center of Quality and Applied Statistics, Rochester Institute of Technology}
\address{School of Mathematical Sciences, Rochester Institute of Technology}

\ead[url]{bl3267@rit.edu, epfeqa@rit.edu}

\address[mymainaddress]{98 Lomb Memorial Drive, Rochester, New York}

\begin{abstract}
We introduce and develop a novel approach to outlier detection based on adaptation of random subspace learning. Our proposed method handles both high-dimension low-sample size and traditional low-dimensional high-sample size datasets. Essentially, we avoid the computational bottleneck of techniques like minimum covariance determinant (MCD) by computing the needed determinants and associated measures in much lower dimensional subspaces. Both theoretical and computational development of our approach reveal that it is computationally more efficient than the regularized methods in high-dimensional low-sample size, and often competes favorably with existing methods as  far as the percentage of correct outlier detection is concerned.
\end{abstract}

\begin{keyword}
\texttt{High-dimensional,} \texttt{Robust,} \texttt{Outlier Detection,} \texttt{Contamination,}
\texttt{Large $p$ small $n$,} \texttt{Random Subspace Method,} \texttt{Minimum Covariance Determinant}
\MSC[2015] 00-01\sep  99-00
\end{keyword}

\end{frontmatter}

\section{Introduction}
We are given a dataset $\mathscr{D} = \{\vx_1,\cdots,\vx_n\}$, where $\vx_i=\left(x_{i1}, \cdots, x_{ip}\right)^{\top} \in \mathscr{X} \subset \mathbb{R}^{1\times p}$, under the special scenario in which $n\lll p$, referred to as high dimensional low sample size (HDLSS) setting.
It is assumed that the basic distribution of the $X_i$'s is multivariate Gaussian, so that the density of $X$ is given by $\phi_p(\vx; \bfmu, \Sigma)$, with
         \begin{eqnarray}
             \label{eq:p:gauss}
             \phi_p(\vx; \bfmu, \Sigma) = \frac{1}{\sqrt{(2\pi)^p |\bfSigma|}} \exp\left\{-\frac{1}{2}(\vx-\bfmu)^\top \bfSigma^{-1}(\vx-\bfmu)\right\}.
         \end{eqnarray}
It is also further assumed that the data set $\mathscr{D}$ is contaminated, with a proportion $\varepsilon \in (0, \tau)$ where $\tau < e^{-1}$, of observations that are outliers, so that under $\varepsilon$-contamination regime, the probability density function of $X$ is given by
         \begin{eqnarray}
            \label{eq:2:1a}
            p(\vx | \bfmu, \bfSigma, \varepsilon, \eta, \gamma) =
           (1-\varepsilon)\phi_p(\vx; \bfmu,\bfSigma)+\varepsilon \phi_p(\vx; \bfmu+\eta, \gamma\bfSigma),
         \end{eqnarray}
         where $\eta$ represents the contamination of the location parameter $\bfmu$, while $\gamma$ captures the
         level of contamination of the scatter matrix $\bfSigma$.
Given a dataset with the above characteristics, the goal of all outlier detection techniques and methods is to {\it select and isolate as many outliers as possible so as to perform robust statistical procedures non-aversely affected by those outliers.}
In such scenarios where the multivariate Gaussian is the assumed basic underlying distribution, the
classical Mahalanobis distance is the default measure of the proximity of the observations, namely
\begin{eqnarray}
    \label{eq:maha:1}
    d_{\bfmu, \bfSigma}^2(\vx_i)=\left(\vx_i-\bfmu\right)^{\top}\bfSigma^{-1}\left(\vx_i-\bfmu\right),
\end{eqnarray}
and experimenters of often address and tackle the outlier detection task in such situations
using either the so-called Minimum Covariance Determinant (MCD) Algorithm \cite{Rousseeuw:1984:1} or some extensions or adaptations
thereof.

\begin{algorithm}[H]
\caption{Minimum Covariance Determinant (MCD)}\label{algo:mcd:1}
\begin{algorithmic}[1]
\State Select $h$ observations, and form the dataset $\mathscr{D}_H$. $H\subset \{1,\cdots,n\}.$ 
\State Compute the empirical covariance $\widehat{\bfSigma}_{H}$ and mean $\widehat{\bfmu}_{H}$.
\State Compute the Mahalanobis distances
$\,\, d_{\widehat{\bfmu}_H, \widehat{\bfSigma}_H}^2(\vx_i), \,\,\,i=1,\cdots,n$
\State Select the $h$ observations having the smallest Mahalanobis distance.
\State Update $\mathscr{D}_{H}$ and repeat steps $2$ to $5$ until ${\tt det}(\widehat{\bfSigma}_H)$ no longer decreases.
\label{algo:mcd:1}
\end{algorithmic}
\end{algorithm}
The  MCD algorithm can be formulated as an optimization problem
$$
(\widehat{H}, \widehat{\bfmu}_H, \widehat{\bfSigma}_H) = \underset{\bfmu, \bfSigma, H}{\tt argmin}\left\{\mathcal{E}(\bfmu, \bfSigma, H)\right\}
$$
where
$$
\mathcal{E}(\bfmu, \bfSigma, H) = \log\{{\tt det}(\bfSigma)\} +
\frac{1}{h}\sum_{i \in H}{\left(\vx_i-\bfmu\right)^{\top}\bfSigma^{-1}\left(\vx_i-\bfmu\right)}.
$$
The seminal MCD algorithm proposed by \cite{Rousseeuw:1984:1} turned out to be rather slow and did not scale well as a function of the sample size $n$. That limitation of MCD led its author to creation of the so-called FAST-MCD \cite{Rousseeuw:1999:1}, focused on solving the outlier detection problem in a more computationally efficient way. Since the algorithm only needs to select a limited number $h$  of observations for each loop, its complexity can be reduced when sample size $n$ is large, since only a small fraction of the data is used. It must be noted however that the bulk of the computations in MCD has to do with the estimation of determinants and the Mahalanobis distances, both requiring a complexity of $O(p^3)$ where $p$ is the dimensionality of the input space as defined earlier. It becomes crucial therefore to find out how MCD fares when $n$ is large and $p$ is also large, even the now quite ubiquitous scenario where $n$ is small but  $p$ is very larger, and indeed much larger than $n$. This {\it $p$ larger than $n$} scenario, referred to as high dimension low sample size (HDLSS) is very common nowadays in application domains such as gene expression datasets from RNA-sequencing and microarray, audio processing, image processing, just to name a few. As noted before, with the MCD algorithm, $h$ observations have to be selected to compute the robust estimator. Unfortunately, when $n\lll p$, neither the inverse nor the determinant of covariance matrix can be computed. As we'll show later, the $O(p^3)$ complexity of matrix inversion and determinant computatation renders MCD untenable for $p$ as moderate as $500$. It is therefore natural, in the presence of  HDLSS datasets, to contemplate at least some intermediate dimensionality reduction step prior to performing the outlier detection task. Several algorithms have been proposed, among which PCOut by \cite{Filzmoser20081694}, Regularized MCD (R-MCD) by \cite{Fritsch_2011} and other ideas by \cite{angiulli2002fast}, \cite{Aggarwal:2005:EEA:1057490.1057495}, \cite{ghoting2008fast}, \cite{kriegel2009outlier}.
When instability in the data makes the computation of $\widehat{\bfSigma}$ problematic in $p$ dimension, regularized MCD may be used with objective function
$$
 \mathcal{E}(\bfmu, \bfSigma, H, \lambda) = \mathcal{E}(\bfmu, \bfSigma, H) + \lambda {\tt trace}(\bfSigma^{-1}),
$$
where $\lambda$ is the so-called regularizer or tuning parameter, chosen to stabilize the procedure.
However, it turns out that even the above Regularized MCD cannot be contemplated when $p \ggg n$, since ${\tt det}(\widehat{\bfSigma})$ is always zero in such cases. The solution to that added difficulty is addressed by solving
\begin{eqnarray}
\left(\widehat{H}, \widehat{\bfmu}_H,\widehat{\bfSigma}_H\right)={\tt arg}{\tt max}{\Big\{\log\{{\tt det}(\widetilde{\bfSigma})\} +
\frac{1}{h}\sum_{i \in H}{\left(\vx_i-\bfmu\right)^{\top}\widetilde{\bfSigma}^{-1}\left(\vx_i-\bfmu\right)}+\lambda{\mathtt{trace}(\widetilde{\bfSigma}^{-1})}\Big\}}
\end{eqnarray}
where the regularized coveriance matrix $\widetilde{\bfSigma}$ is given by
$$
 \tilde{\bfSigma}(\alpha) = (1-\alpha)\widehat{\bfSigma} + \frac{\alpha}{p}{\tt trace}(\widehat{\bfSigma})\bfI_p
$$
with $\alpha \in (0,1)$.  For many HDLSS datasets however, the dimensionality $p$ of the input space is often large, with numbers like  $p\geq10^{3}$ or even $p\geq10^{4}$ rather very common. As a result, even the above direct regularization is computationally intractable, because when $p$ is large, the $O(p^3)$ complexity of the needed matrix inversion and determinant calculation makes the problem computationally untenable. The fastest matrix inversion algorithms like \cite{Coppersmith1990251} and \cite{LeGall:2014:PTF:2608628.2608664} are theoretically around $O(p^{2.376})$ and $O(p^{2.373})$, and so complicated that there are virtually no useful implementation of any of them. In short, the regularization approach to MCD like algorithms is impractical and unusable for HDLSS datasets even for values of $p$ around a few hundreds.
Another approach to outlier detection in the HDLSS context has revolved around extensions and adaptations of principle component analysis(PCA).
Classical PCA seeks to project high dimensional vectors onto a lower dimensional orthogonal space while maximizing the variance. By reducing the dimensionality of the original data, one seeks to create a new data representation that evades the curse of dimensionality. However, PCA, in its
generic form, is not robust, for the obvious reason that it is built by a series of transformations of means and covariance matrices whose generic estimators are notoriously non robust. It is therefore of interest to seek to perform PCA in a way that does not suffer from the
 presence of outliers in the data, and thereby identify the outlying observations as a byproduct of such a PCA. Many authors have worked on the robustification of PCA, and among them \cite{Hubert:2004:1} whose proposed ROBPCA, a robust PCA method, which  essentially robustifies PCA by combining MCD with the famous {\it projection pursuit} technique (\cite{croux1996fast}, \cite{citeulike:6057047}). Interestingly, if instead of reducing the dimensionality based on robust estimators, one can first apply PCA to the whole data, then outliers may surprisingly lie on several directions where they are then exposed more clearly and distinctly. Such an insight appear to have motivated the creation of the so-called PCOut algorithm proposed by \cite{Filzmoser20081694}.  PCOut uses PCA as part of its preprocessing step after the original data has been scaled by Median Absolute Deviation (MAD). In fact, in PCOut, each attribute is transformed as follows:
\begin{eqnarray}
\vx_j^{\ast} = \frac{\vx_j - \widetilde{\vx}_j}{MAD(\vx_j)}, j=1,\cdots,p,
\end{eqnarray}
where $\vx_j = \left(x_{1j}, \cdots, x_{nj}\right) \subset \mathbb{R}^{n\times 1}$ and $\widetilde{\vx}_j$ is the median of $\vx_j$. Then with ${\boldsymbol{X}}^{\ast} = \big[\vx_1^{\ast},\vx_2^{\ast},\cdots,\vx_p^{\ast}\big]$, PCA can be performed, namely
\begin{eqnarray}
{{\boldsymbol{X}}^{\ast}}^\top {\boldsymbol{X}}^{\ast} = \bfV \bfLambda \bfV^\top
\end{eqnarray}
from which the principal component scores $\boldsymbol{Z} = {\boldsymbol{X}}^{\ast}\cdot\boldsymbol{V}$ may then be used for the purpose of
outlier detection. In fact, it also turns out that the principal component scores $\boldsymbol{Z}$ may be re-scaled to achieve a much lower dimension with $99\%$ variance retained. Unlike MCD, PCA based re-scaled method is not only practical but also performs better with high dimensional datasets. $99\%$ of simulated outliers are detected when $n = 2000, p = 2000$. A higher false positive rate is reported in low dimensional cases, and less than half of the outliers were identified in scenarios with $n = 2000, p = 50$.
It is clear by now that with HDLSS datasets, some form of dimensionality reduction is needed prior to performing outlier detection. Unlike the authors just mentioned who all resorted to some extension or adaptation of principal component analysis wherein dimensionality reduction is based on transformational projection, we herein propose an approach where dimensionality reduction is not only stochastic but also selection-based rather than projection-based. The rest of this paper is organized as follows: in section 2, we present a detailed description of our proposed approach, along with all the needed theoretical and conceptual justifications.  In the interest of completeness, we close this section with the general description of a nonparametric machine learning kernel method for novelty detection known as the one-class support vector machine, which under suitable conditions is an alternative to the outlier detection approach proposed in this paper. Section 3 contains our extensive computational demonstrations on various scenarios. We specifically present the comparisons of the predictive/detection  performances between our RSSL based approach and the PCA based methods discussed earlier. We mainly used simulated data here, with simulations seeking to assess the impact of various aspects of the data such as the dimensionality $p$ of the input space, the contamination rate $\varepsilon$ and other aspects like the magnitude $\gamma$ of the contamination of the scatter matrix. We conclude with section 4, in which we provide a thorough discussion of our results along with various pointers to our current and future work on this rather compelling theme of outlier detection.

\section{Random Subspace Learning Approach to Outlier Detection}
\subsection{Rationale for Random Subspace Learning}
We herein propose a technique that combines the concept underlying Random Subspace Learning (RSSL) by \cite{ho1998random} with some of the key ideas behind minimum covariance determinant (MCD) to achieve a computational efficient, scalable, intuitive appealing and highly accurate outlier detection method for both HDLSS and LDHSS datasets. With our proposed method, the computation of the robust estimators of both location and scatter matrix can be achieved by tracing the optimal subspaces directly. Besides, we demonstrate via practical examples that our RSSL based method  is computationally very efficient, specifically because it turns out that, unlike the other methods mentioned earlier, our method does not require the computationally expensive calculations of determinants  and Mahalanobis distances at each step. Morever, whenever such calculations are needed, they are all performed in very low dimensional spaces, further emphasizing the computational strength of our approach.
The original MCD algorithm formulates the outlier detection problem as the problem of finding the smallest determinant of covariances computed from a sequence $\mathscr{D}_{h}^{(k)},\,\, k=1,\cdots,m$ of different subsets of the original data set $\mathscr{D}$. Each subset contains $h$ observations. More precisely, if $\mathscr{D}_{optimal}$ is the subset of $\mathscr{D}$ whose observations yield the estimated covariance matrix with the smallest (minimum) determinant out of all the $m$ subsets considered, then we must have
$$
{\tt det}(\widehat{\bfSigma}(\mathscr{D}_{optimal})) = {\min}\left\{{\tt det}(\widehat{\bfSigma}(\mathscr{D}^{(1)}_h)), {\tt det}(\widehat{\bfSigma}(\mathscr{D}^{(2)}_h)),\cdots,{\tt det}(\widehat{\bfSigma}(\mathscr{D}^{(m)}_h))\right\},
$$
where $m$ is the number of iterations needed for the MCD algorithm to converge. $\mathscr{D}_{optimal}$ is the subset of $\mathscr{D}$ that produces the estimated covariance matrix with the smallest determinant. The MCD estimates of the location vector and scatter matrix parameters are given by
$$
\widehat{\bfmu}_{\tt MCD} = \widehat{\bfmu}(\mathscr{D}_{optimal}) \quad \text{and} \quad \widehat{\bfSigma}_{\tt MCD} = \widehat{\bfSigma}(\mathscr{D}_{optimal}).
$$
The number $h$ of observations in each subset is required to be $\frac{n}{2} \leq h < n$. It turns out that $h=[(n+p+1)/2]$ reaches its highest possible breakdown value according to \cite{lopuhaa1991}. It is obvious that with $h=[(n+p+1)/2]$ being the highest breakdown point, the requirement $\frac{n}{2} \leq h < n$ cannot achieved in the HDLSS context, since in such a context $p \ggg n$. It is therefore intuitively appealing to contemplate a subspace of the input space $\mathscr{X}$, and define/contruct such a subspace in such a way that its dimensionality $d < p$ is also such that $d < n$ to allow the seamless computation of the needed distances.

\subsection{Description Random Subspace Learning for Outlier Detection}
Random Subspace Learning in its generic form is designed for precisely this kind of procedure. In a nutshell, RSSL combines instance-bagging (bootstrap ie sampling observations with replacement) with attribute-bagging (sampling indices of attributes without replacement), to allow efficient ensemble learning in high dimensional spaces.
Random Subspace Learning (Attribute Bagging) proceeds very much like traditional bagging, with the  added crucial step consisting of selecting a subset of the variables from the input space for training rather than building each base learners using all the $p$ original variables.
\begin{algorithm}[H]
\caption{Random Subspace Learning (RSSL): Attribute-bagging step}\label{algo:rssl:1}
\begin{algorithmic}[1]
   \State {\tt Randomly draw the number $d<p$ of variables to consider}
   \State {\tt Draw without replacement the indices of $d$ variables of the original $p$ variables} 
   \State {\tt Perform learning/estimation in the $d$-dimensional subspace}
\label{algo:rssl:1}
\end{algorithmic}
\end{algorithm}
This attribute-bagging step is the main ingredient of our outlier detection approach in high dimensional spaces.

\begin{algorithm}[H]
\caption{Random Subspace Learning for Outlier Detection when $p \lll n$}\label{algo:rs:1}
\begin{algorithmic}[1]
\Procedure{Random Subspace Outlier}{$B$}
\For{$b=1$ to $B$}
\State {\tt Draw with replacement} $\{i_1^{(b)},\cdots,i_n^{(b)}\}$ from $\{1,2,\cdots,n\}$ to form the bootstrap sample $\sD^{(b)}$
\State {\tt Draw without replacement} from $\{1,2,\cdots,p\}$ a subset $\{j_1^{(b)},\cdots,j_d^{(b)}\}$ of $d$ variables
\State {\tt Drop unselected variables} from $\sD^{(b)}$ so that $\sD_{sub}^{(b)}$ is $d$ dimensional
\State {\tt Build the $b$th determinant of covariance} ${\tt det}(\widehat{\bfSigma}(\sD_{sub}^{(b)}))$
\EndFor \label{algo:rs:1}
\State {\tt Sort the ensemble} $\Big\{{\tt det}(\widehat{\bfSigma}(\sD_{sub}^{(b)})),\, b=1,\cdots, B\Big\}$
\State {\tt Form $\sD^{\ast}:{\tt det}(\sD^{\ast}) = {\tt arg}{\tt min}\Big\{{\tt det}(\widehat{\bfSigma}(\sD_{sub}^{(b)})),\, b=1,\cdots, B\Big\}$}
\State {\tt Compute $\widehat{\mu}^{\ast}$ and $\widehat{\bfSigma}^{\ast}$ base on $\sD^{\ast}$}
\State We can build the robust distance by:
\begin{eqnarray}
\widehat{\delta^{\ast}}(\vx) = \left(\vx - \widehat{\mu}^{\ast} \right)^\top {\widehat{\bfSigma}}^{\ast^-1} \left(\vx - \widehat{\mu}^{\ast} \right).
\end{eqnarray}
\EndProcedure
\end{algorithmic}
\end{algorithm}
The RSSL outlier detection algorithm computes a determinant of covariance for each subsample, with each subsample residing in a subspace spanned by the $d$ randomly selected variables, where $d$ is usually selected to be $\min(\frac{n}{5}$, $\sqrt{p})$. A total of $B$ subsets are generated, and their low dimensional covariance matrices are formed along with the corresponding determinants. Then the best subsample, meaning the one with the smallest covariance determinant is singled. It turns out that in the LDHSS context ($n \ggg p $),  our RSSL outlier detection algorithm always robustly yields the robust estimators $\widehat{\mu}^{\ast}$ and $\widehat{\bfSigma}^{\ast}$ needed to compute the Mahalanobis distance for all the observations. Then the outliers can be selected using the typical cut-off built on classical $\chi_{p, 5\%}^2$. In HDLSS context, in order to handle the curse of dimensionality, we need to involve a new variable selection procedure to adjust our framework and concurrently stabilize the detection. The modified version of our RSSL outlier detection algorithm in HDLSS is then given by:

\begin{algorithm}[H]
\caption{Random Subspace Learning for Outlier Detection when $n \lll p$}\label{algo:rf:1}
\begin{algorithmic}[2]
\Procedure{Random Subspace Determinant Covariance}{$B$}
\For{$b=1$ to $B$}
\State {\tt Draw with replacement} $\{i_1^{(b)},\cdots,i_n^{(b)}\}$ from $\{1,2,\cdots,n\}$ to form the bootstrap sample $\sD^{(b)}$
\State {\tt Draw without replacement} from $\{1,2,\cdots,p\}$ a subset $\{j_1^{(b)},\cdots,j_d^{(b)}\}$ of $d$ variables
\State {\tt Drop unselected variables} from $\sD^{(b)}$ so that $\sD_{sub}^{(b)}$ is $d$ dimensional
\State {\tt Build the $b$th determinant of covariance} ${\tt det}(\widehat{\bfSigma}(\sD_{sub}^{(b)}))$
\EndFor \label{algo:rf:2}
\State {\tt Sort the ensemble} $\Big\{{\tt det}(\widehat{\bfSigma}(\sD_{sub}^{(b)})),\, b=1,\cdots, B\Big\}$
\State {\tt Keep the $k$ smallest samples based on elbow to form $\sD^{(\eta)}$, where $\eta=1,\cdots,k$, $k < B$}
\For{$j=2$ to $d$}
\State {\tt Select $\nu=j$ most frequent variables left in $\sD^{(\eta)}$ to compute ${\tt det}(\widehat{\bfSigma}(\sD_{sub=j}^{(\eta=1)}))$}
\EndFor
\State {\tt Form $\sD^{\ast}:{\tt det}(\sD^{\ast}) = {\tt arg}{\tt max}\Big\{{\tt det}(\widehat{\bfSigma}(\sD_{sub=j}^{(\eta=1)})),\, j=2,\cdots, d\Big\}$}
\State {\tt Compute $\widehat{\mu}^{\ast}$ and $\widehat{\bfSigma}^{\ast}$ base on $\sD^{\ast}$}
\State We can build the robust distance by:
\begin{eqnarray*}
\widehat{\delta^{\ast}}(\vx) = \left(\vx - \widehat{\mu}^{\ast} \right)^\top {\widehat{\bfSigma}}^{\ast^-1} \left(\vx - \widehat{\mu}^{\ast} \right).
\end{eqnarray*}
\EndProcedure
\end{algorithmic}
\end{algorithm}
Without selecting the smallest determinant of covariance, we choose to select a certain number of subsamples to achieve the variable selection through a sort of voting process. The portion of the most frequently appearing variables are elected to build an optimal space that allow us to compute our robust estimators. The simulation results and other details will be discussed later.

\subsection{Justification Random Subspace Learning for Outlier Detection}
\begin{conjecture}
Let $\sD$ be the dataset under consideration. Assume that a proportion $\varepsilon$ of the observations in $\sD$ are
outliers. If $\varepsilon < e^{-1}$, then will high probability, the proposed RSSL outlier detection algorithm will efficiently
correctly identify a set of data that contains very few of the outliers.
\end{conjecture}
\begin{sketch}
Let $\vx_i \in \sD$ be a random observation in the original dataset $\sD$. Let $\sD^{(b)}$ denote the $b$th bootstrapped sample from $\sD$. Let $\Pr[\vx_i \in \sD^{(b)}]$ represent the proportion of observations that are in $\sD$ but also present in $\sD^{(b)}$. It is easy to prove
$\Pr[\vx_i \in \sD^{(b)}]
                         = 1- \left(1-\frac{1}{n}\right)^n$.
In other words, if $\Pr[\vx_i \notin \sD^{(b)}]=\Pr[O_n]$ denotes the observations from $\sD$ not present in $\sD^{(b)}$, we must have
$\Pr[\vx_i \notin \sD^{(b)}]=\left(1-\frac{1}{n}\right)^n = \Pr[O_n]$.
Since $\Pr[O_n]$ is known to converge to $e^{-1}$ as $n$ goes to infinity.  Therefore for each given bootstrapped sample $\sD^{(b)}$, there is
a probability close to $e^{-1}$ that any given outlier will not corrupt the estimation of location vector and scatter matrix parameters.
Since the outliers as well as all other observations have an asymptotic probability of $e^{-1}$ of not affecting the bootstrapped estimator that we build. Therefore over a large enough re-sampling process (large $B$), there will be many bootstrapped samples $\sD^{(b)}$ with very few outliers leading to a sequence of small covariance determinants as desired, if $\varepsilon < e^{-1}$. It is therefore reasonable to deduce that by averaging this exclusion of outliers over many replications, robust estimators will naturally be generated by the RSSL algorithm
\end{sketch}

\subsection{Alternatives to Parametric Outlier Detection Methods}
The assumption of multivariate Gaussianity of the $\vx_i$'s is obviously limiting as it could happen that the data does not follow a Gaussian distribution. Outside of the realm where location and scatter matrix play a central role, other methods have been proposed, especially in the field of machine learning, and specifically with similarity measures known as kernels.  One such method is known as One-Class Support Vector Machine (OCSVM) proposed by \cite{Scholkopf99estimatingthe} to solve the so-called novelty detection problem. It is important to emphasize right away that novelty detection although similar in spirit to outlier detection, can be quite different when it comes to the way the algorithms are trained.
OCSVM approach to novelty detection is interesting to mention here because despite some conceptual differences from the covariance methods explored earlier, it is formidable at handling HDLSS data thanks to the power of kernels. Let $\Phi: \mathscr{X} \longrightarrow \mathscr{F}$. The one-class SVM novelty detection solves
               $$
               \underset{\vw\in \mathscr{F},\bfxi \in \Real^n,\rho\in \Real}{\tt argmin}\left\{ \frac{1}{2}\|\vw\|^2 + \frac{1}{\nu n}\sum_{i=1}^n{\xi_i} -\rho\right\}
               $$
               Subject to
               $$
               \langle \vw, \Phi(\vx_i)\rangle > \rho -\xi_i, \,\,\, \xi_i \geq 0, \,\,\, i=1,\cdots, n
               $$
               Using $\mathscr{K}(\vx_i, \vx_j) = \langle \Phi(\vx_i), \Phi(\vx_j) \rangle = \Phi(\vx_i)^\top \Phi(\vx_j)$, we get
               $$
               \widehat{f}(\vx_i) = {\tt sign}\left(\sum_{j=1}^n{\widehat{\alpha}_j \mathscr{K}(\vx_i,\vx_j) - \widehat{\rho}}\right)
               $$
               so that any $\vx_i$ with $\widehat{f}(\vx_i)<0$ is declared an outlier.
 The $\widehat{\alpha}_j$'s and $\widehat{\rho}$ are determined by solving the quadratic programming problem formulated above
 The parameter $\nu$ controls the proportion of outliers detected.
 One of the most common kernel is the so-called RBF kernel defined by
               $$
               \mathscr{K}(\vx_i,\vx_j) = \exp\left\{-\frac{1}{2\sigma^2}\|\vx_i-\vx_j\|^2\right\}
               $$
OCSVM has been extensively studied and applied  by many researchers  among which \cite{manevitz2002one}, \cite{hubert2005robpca} and \cite{Zhang:2007:OCS:1503549.1503556}, and later enhanced by \cite{Amer:2013:EOS:2500853.2500857}. OCSVM is often applied to semi-supervised learning tasks where training focuses on all the positive examples (non outliers) and then the detection of anomalies is performed by searching points that fall geometrically outside of the estimated/learned decision boundary of the good (non outlying trained instances). It is a concrete and quite popular algorithm for solving one-class problems in fields like digital recognition and documentation categorization. However, it is crucial to note that OCSVM cannot be used with many other real life datasets for which outliers are not well-defined and/or for which there are no clearly identified all-positive training examples available such as gene expression mentioned before.

\section{Computational Demonstrations}
\subsection{Setup of Computational demonstration and initial results}
In this section, we conduct a simulation study to assess the performance of our algorithm based on various important aspects of the data, and we also provide a comparison of the predictive/detection performance of our method against existing approaches. All our simulated data are generated according to the $\varepsilon$-contaminated multivariate Gaussian
introduced via    \eqref{eq:p:gauss} and    \eqref{eq:2:1a}. In order to assess the effect the covariance between the attributes,
we use an AR-type covariance matrix of the following form:
\begin{eqnarray}
\label{eq:ar:sigma}
\Sigma = \left(
                        \begin{array}{ccccc}
                          1 & \rho & \cdots & \cdots &\rho \\
                          \rho & 1 & \rho & \cdots & \rho \\
                          \vdots & \ddots & \ddots & \ddots &  \vdots\\
                          \rho & \ddots & \rho & 1 &  \rho\\
                          \rho & \cdots & \cdots & \rho & 1 \\
                        \end{array}
          \right) = [(1-\rho)\bfI_p + \rho \bf1_p  \bf1_p^\top],
\end{eqnarray}
where $\bfI_p$ is the $p$-dimensional identity matrix, while $\bf1_p$ is
$p$-dimensional vector of ones. For the remaining parameters, we consider 3 different levels of contamination
$\varepsilon \in \{0.05, 0.1, 0.15\}$, namely mild contamination to strong contamination. The dimensionality $p$ will increase in low-dimensional case as $\{30, 40, 50, 60, 70\}$ and high dimensional case as $\{1000, 2000, 3000, 4000, 5000\}$ and the number of observations are fixed at 1500 and 100. We compare our algorithm to existing PCA based algorithms $\tt{PCOut}$ and $\tt{PCDist}$, both of which are available in ${\tt R}$ within the package called ${\tt rrcovHD}$.

\begin{figure}[H]
\centering
  \epsfig{figure=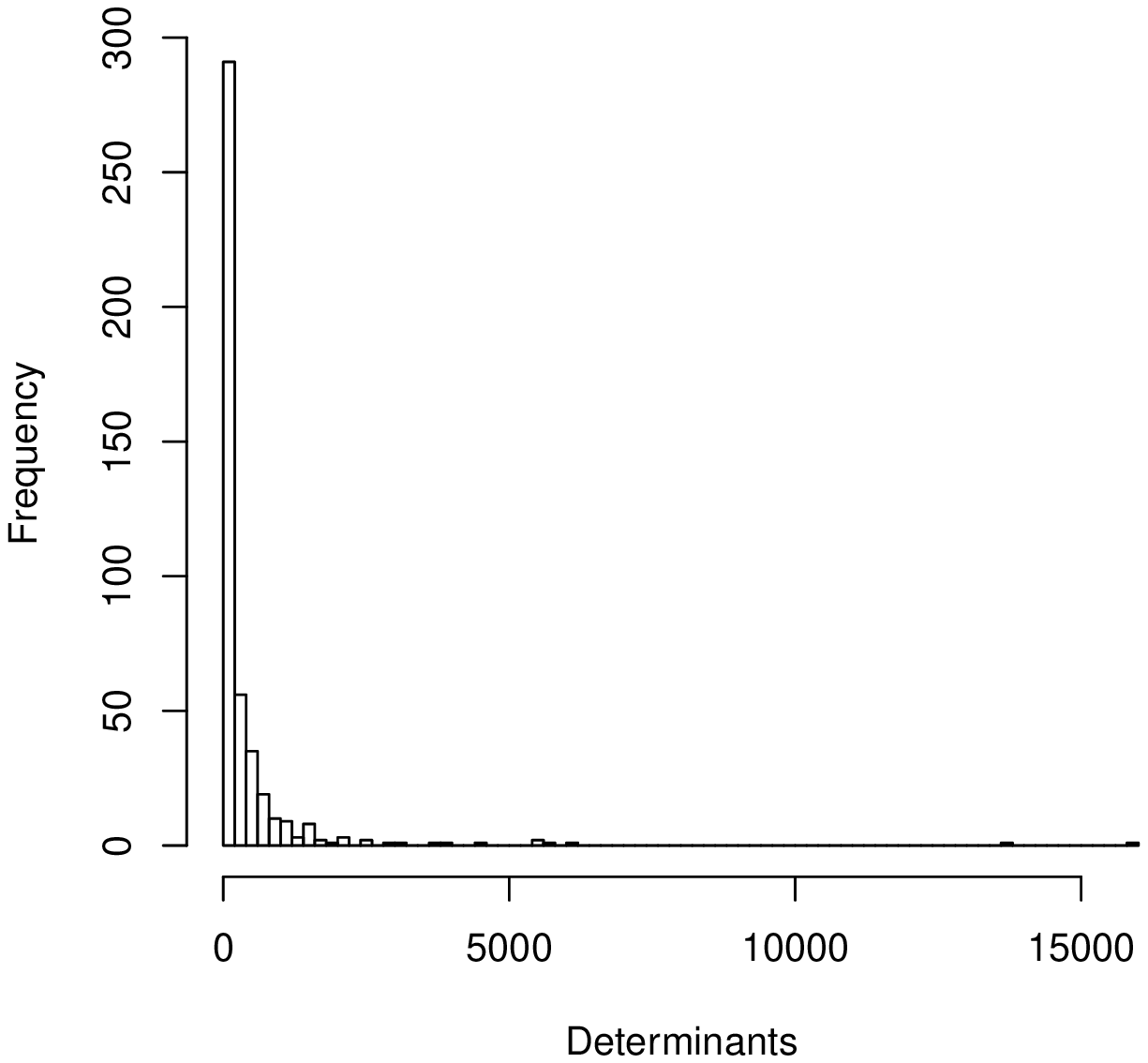, width=7cm,height=7cm}
  \epsfig{figure=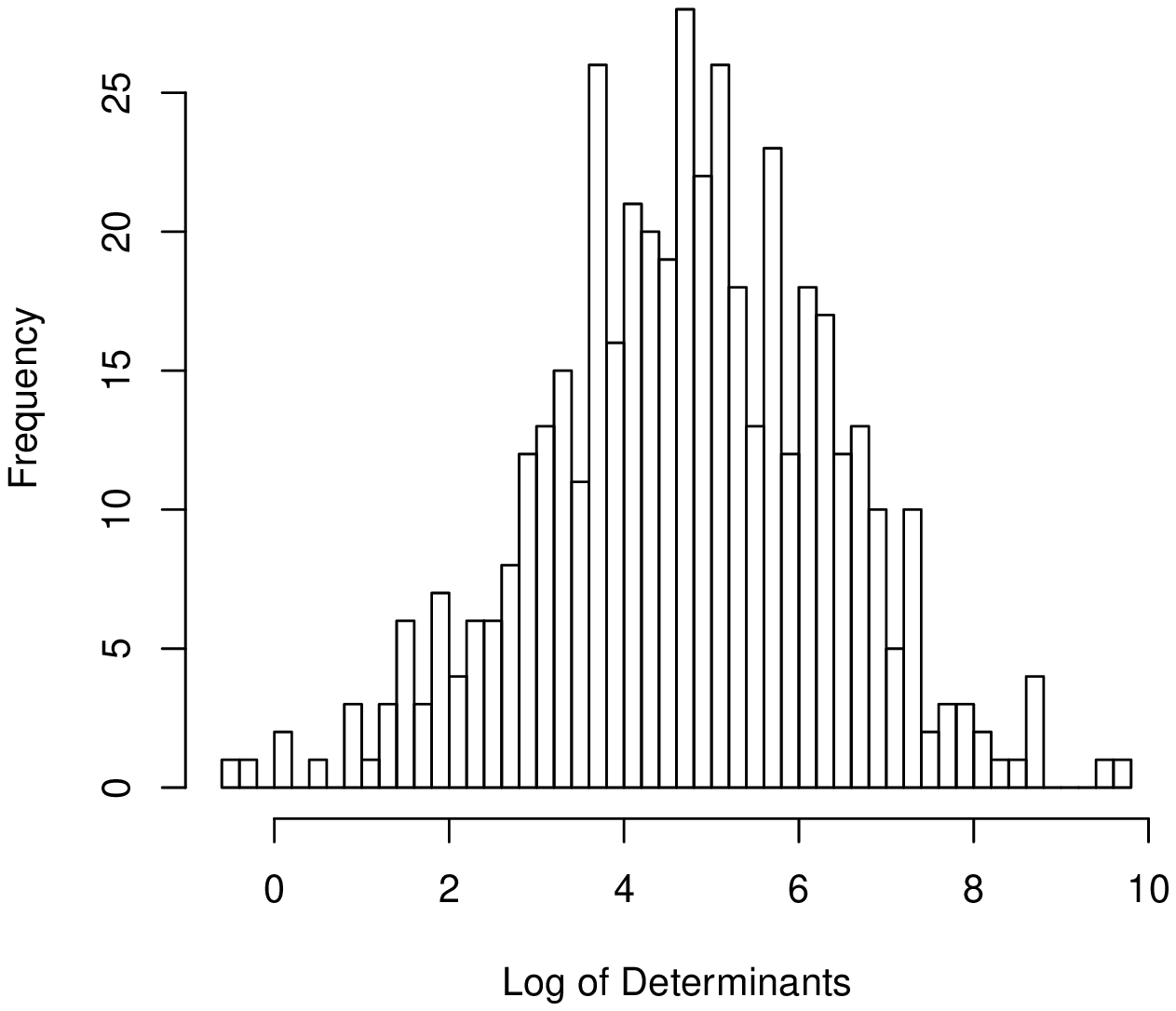, width=7cm,height=7cm}
  \caption{(left) Histogram of the distribution of the determinants from all bootstrap samples $\sD_{\tt sub}^{(b)}$ when $n = 100$, $p = 3000$;
  (right) Histogram of log determinants for all the bootstrap samples. Our methodology later selects a portion of samples based on what we call here the elbow.}
  \label{fig:1}
\end{figure}

As can be seen on Figure \eqref{fig:1}, the overwhelming majority of samples lead to determinants that are small as evidenced by the heavy right skewness with concentration around zero. This further confirms our conjecture that as long as $\varepsilon < e^{-1}$ which is a rather reasonable and easily realized assumption, we should isolate samples with few or no outliers.

\begin{figure}[H]
\centering
  \epsfig{figure=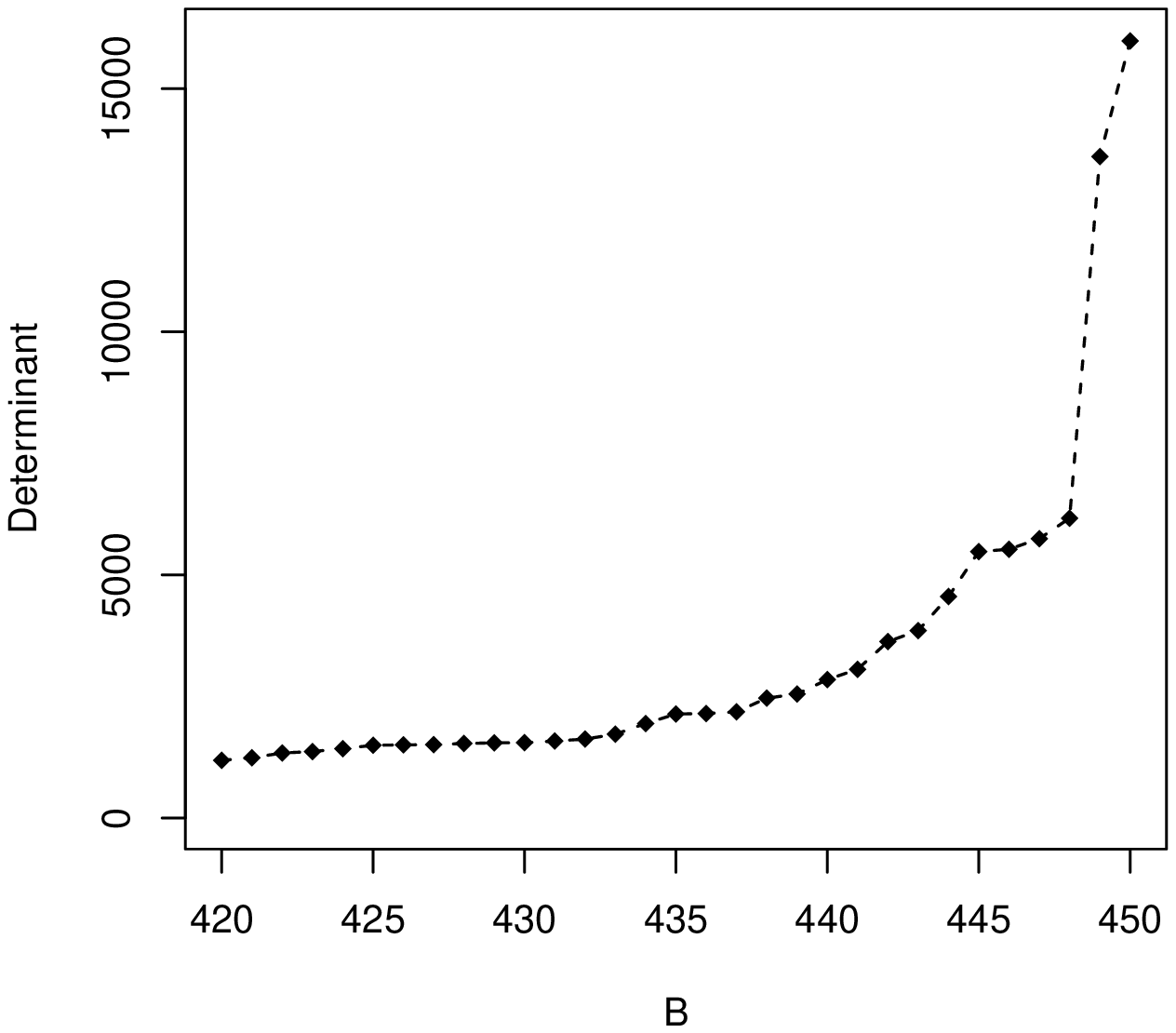, width=7cm,height=7cm}
  \epsfig{figure=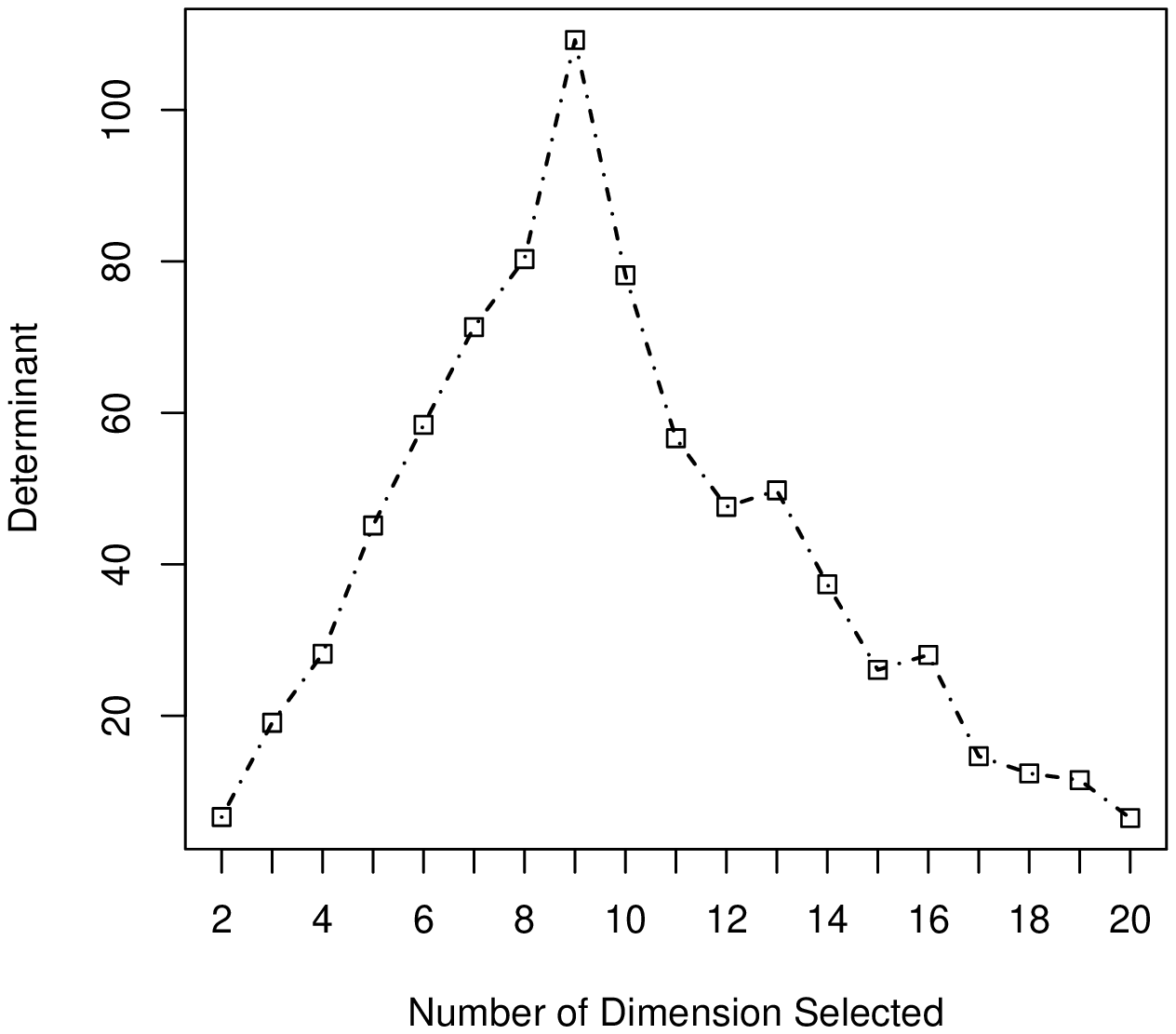, width=7cm,height=7cm}
  \caption{(left) Tail of sorted determinants in high dimensional $\sD_{\tt sub}^{(b)}$, where $B=450$. $k$ can be selected before reaching the elbow;
  (right) The concave shape can be observed by computing determinants of covariance from 2 to $m$ dimension. The cut-off $\nu$ for variable selection is based on the decreasing sorted frequency located at the maximum of the determinants.}
  \label{fig:2}
\end{figure}

Since each bootstrapped sample selected has a small chance of being affected by the outliers, we can select the dimensionality that maximize this benefits. In our HDLSS simulations, determinants are computed based on all the randomly selected subspaces, and are ruled by predominantly small values, which implies the robustness of the classifier. Figure \eqref{fig:1} patently shows the dominance of small values of determinants, which in this case are the determinants of all bootstrapped samples based on our simulated data. A distinguishable elbow is presented in Figure \eqref{fig:2}. The next crucial step lies in selecting a certain number of bootstrap samples, say $k$, to build an optimal subspace. Since most of the determinants are close to each other, it is a non-trivial problem, which means that $k$ needs to be carefully chosen to avoid going beyond the elbow. However, it is important to notice if $k$ is too small then the variable selection in later steps of the algorithm will become a random pick, because there is no opportunity for each variable to appear in the ensemble. Here, we choose $k$ to be the number of roughly the first 30\% to 80\% of $B$ bootstrap samples $\sD^{(\eta)}$ according to their ascending order of the determinants. This choice is based on our empirical experimentations. It is not too difficult to infer the asymptotic normal distribution of the frequencies of all variables in $\sD^{(\eta)}$ as we can observe in Figure \eqref{fig:2}. Thus, the most frequently appearing variables located on the left tail can be adopted/kept to build our robust estimator. Once the selection of $k$ is made, the frequencies of variables appearing in this ensemble can be obtained/computed for variable selection. The 2 to $m$ most frequently appearing variables are included to compute the determinants in Figure \eqref{fig:2}. $m$ is usually small, since we assume from the start that the true dimensionality of the data is indeed small. Here for instance, we choose $20$ for the purposes of our computational demonstration. A sharp maximum indicates the number of dimension $\nu$ from that sorted ensemble that we need to choose. Thus, with the bootstrapped observations having the smallest determinant with the subspace that generates the largest determinant, we can successfully compute $\sD^{\ast} = \sD_{sub=\nu}^{(\eta=1)}$. Then the robust estimators can be formed by $\widehat{\mu}^{\ast}$ and ${\widehat{\bfSigma}}^{\ast}$. Theoretically then we are in a presence of a minimax formulation of our outlier detection problem, namely
\begin{eqnarray}
\label{eq:minimax:rssl:1}
\{\sD^{(*)}, \mathscr{V}^{(*)}\} = \underset{\mathscr{V}^{(b)}}{\tt argmax}\left\{\underset{\mathscr{D}^{(b)}}{\tt argmin}\{{\tt det}({\tt cov}(\hat{\bfSigma}(\mathscr{D}^{(b)}(\mathscr{V}^{(b)}))))\}\right\}
\end{eqnarray}
By Equation \label{eq:minimax:rssl:1}, it should be understood that we need to isolated the precious subsample $\sD^{(*)}$ that achieves the smallest overall covariance determinant, but then concurrently identify along with $\sD^{(*)}$ the subspace $\mathscr{V}^{(*)}$ that yields the highest value of that covariance determinant among all the possible subspaces considered.

\subsection{Further results and Computational Comparisons}
\begin{figure}[H]
\centering
  \epsfig{figure=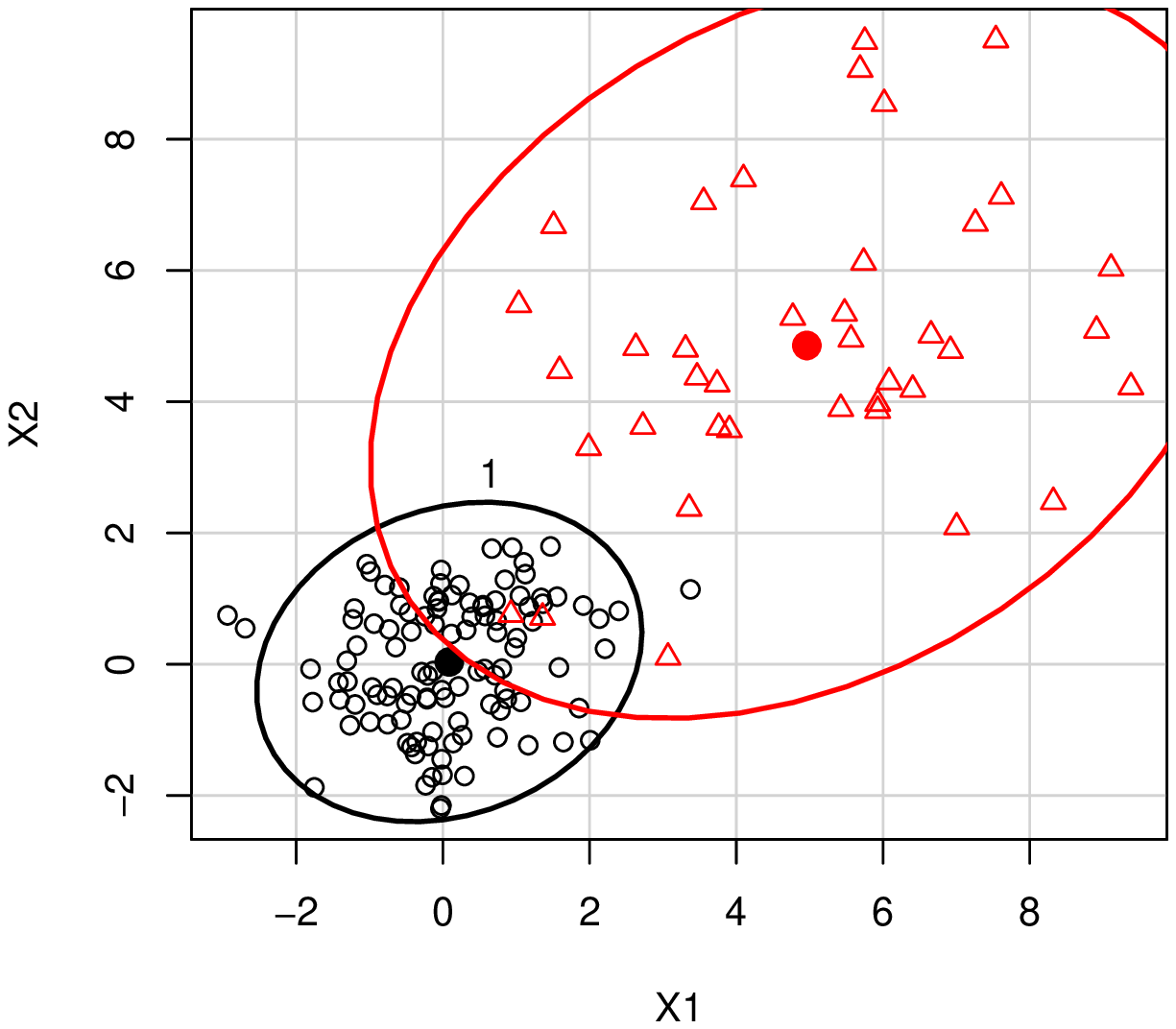, width=7cm,height=7cm}
  \epsfig{figure=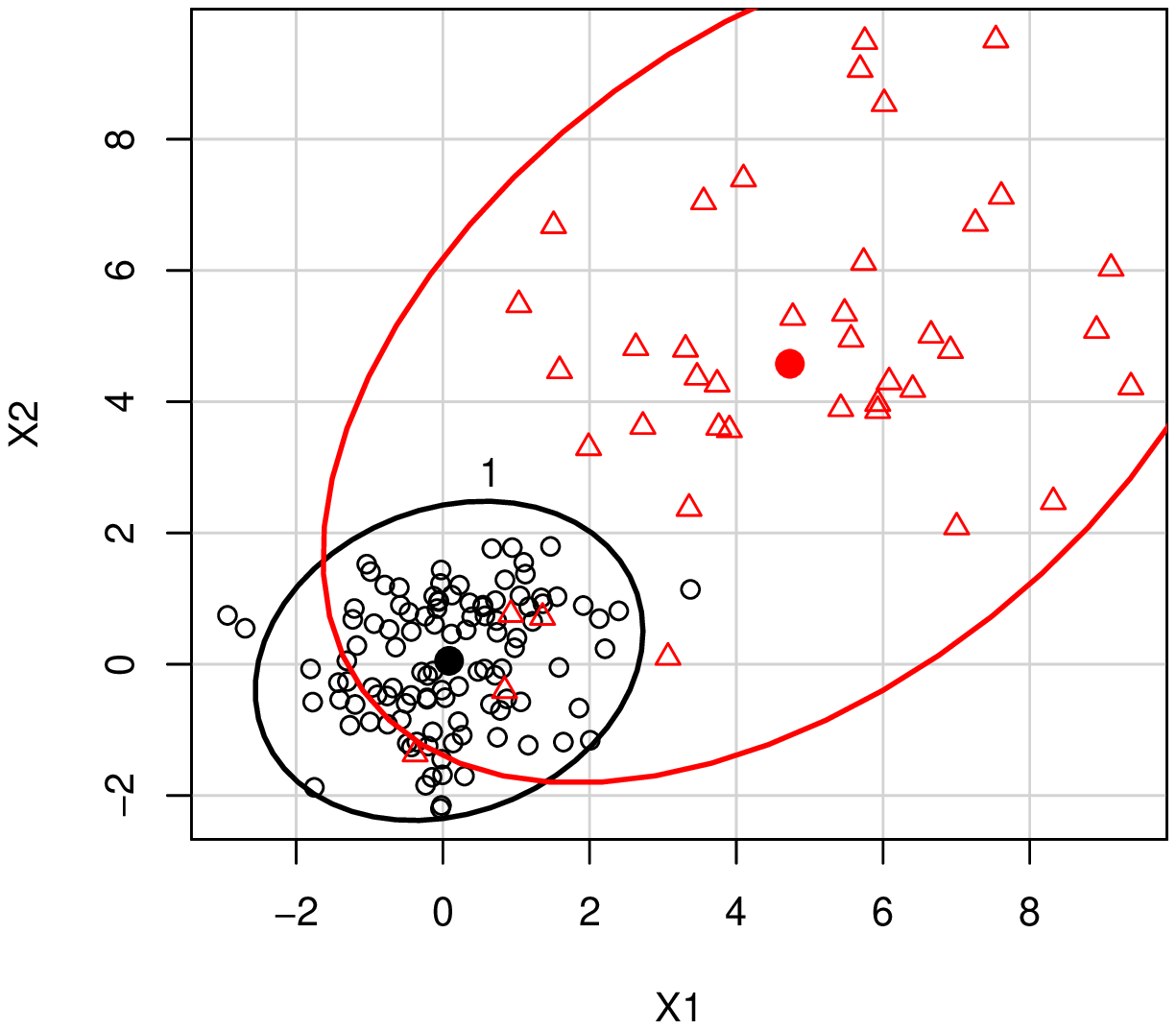, width=7cm,height=7cm}
  \caption{(left) The outliers detected in a two dimensional subspace are marked as red triangles. Selection is based on $\widehat{\delta^{\ast}}(\vx) > \chi_{df = d, \alpha = 5\%}^2$;
  (right) Outliers are selected by $\chi_{df = d, \alpha = 10\%}^2$.}
  \label{fig:3}
\end{figure}

As indicated in our introductory section, we use the Mahalanobis distance as our measure of proximity. As since we are operating under the assumption of multivariate normality, we use the traditional distribution quantiles $\chi_{d, \frac{\alpha}{2}}^2$ as our cut-off with the typical $\alpha = 10\%$ and $\alpha = 5\%$. As usual, all observations with distances larger than $\chi_{d, \frac{\alpha}{2}}^2$ are classified as outliers. The data for simulation study are generated with $\eta, \kappa \in \{2, 5\}$ representing both easy and hard situation for RSSL algorithm to detect the outliers, and $\varepsilon$, $p$ as the rate of contamination. Throughout, we use $R = 200$ replications for each combination of parameters for each algorithm, and we use the average test error $\tt{AVE}$ as our measure of predictive/detection performance. Specifically,
\begin{eqnarray}
\mathtt{AVE}(\widehat{f}) =\frac{1}{R} \sum_{r=1}^{R} \left\{ \frac{1}{m} \sum_{i=1}^{m} \ell (\vy_{i}^{(r)}, \widehat{f}_{r}(\vx_i^{(r)}))\right\},
\end{eqnarray}
where $\widehat{f}_{r}(\vx_i^{(r)}))$ is the predicted label of the test set observation $i$ yielded by $\widehat{f}$ in the $r$-th replication.
The loss function used here is the basic zero-one loss defined by:
\begin{eqnarray}
        \ell(\vy_{i}^{(r)}, \widehat{f}_{r}(\vx_i^{(r)})) = 1_{\{\vy_{i}^{(r)} \neq \widehat{f}_{r}(\vx_i^{(r)})\}}
                                           = \left\{\begin{array}{ll}
                                                 1 & \mbox{if $\vy_{i}^{(r)} \neq \widehat{f}_{r}(\vx_i^{(r)})$}\\
                                                 0 & \mbox{otherwise}.
                                                     \end{array}\right.
\end{eqnarray}
It will be seen later that our proposed method produces predictive accurate outlier detection results, typically competing favorably against other techniques, and usually outperforming them.  Firstly however,  we show  in Figure \eqref{fig:3} the detection performance of our algorithm based on two randomly selected subspaces. The outliers detected by our algorithm are identified by red triangles and contained in the red contour, while the black circles are the normal data.

\begin{figure}[H]
\centering
  \epsfig{figure=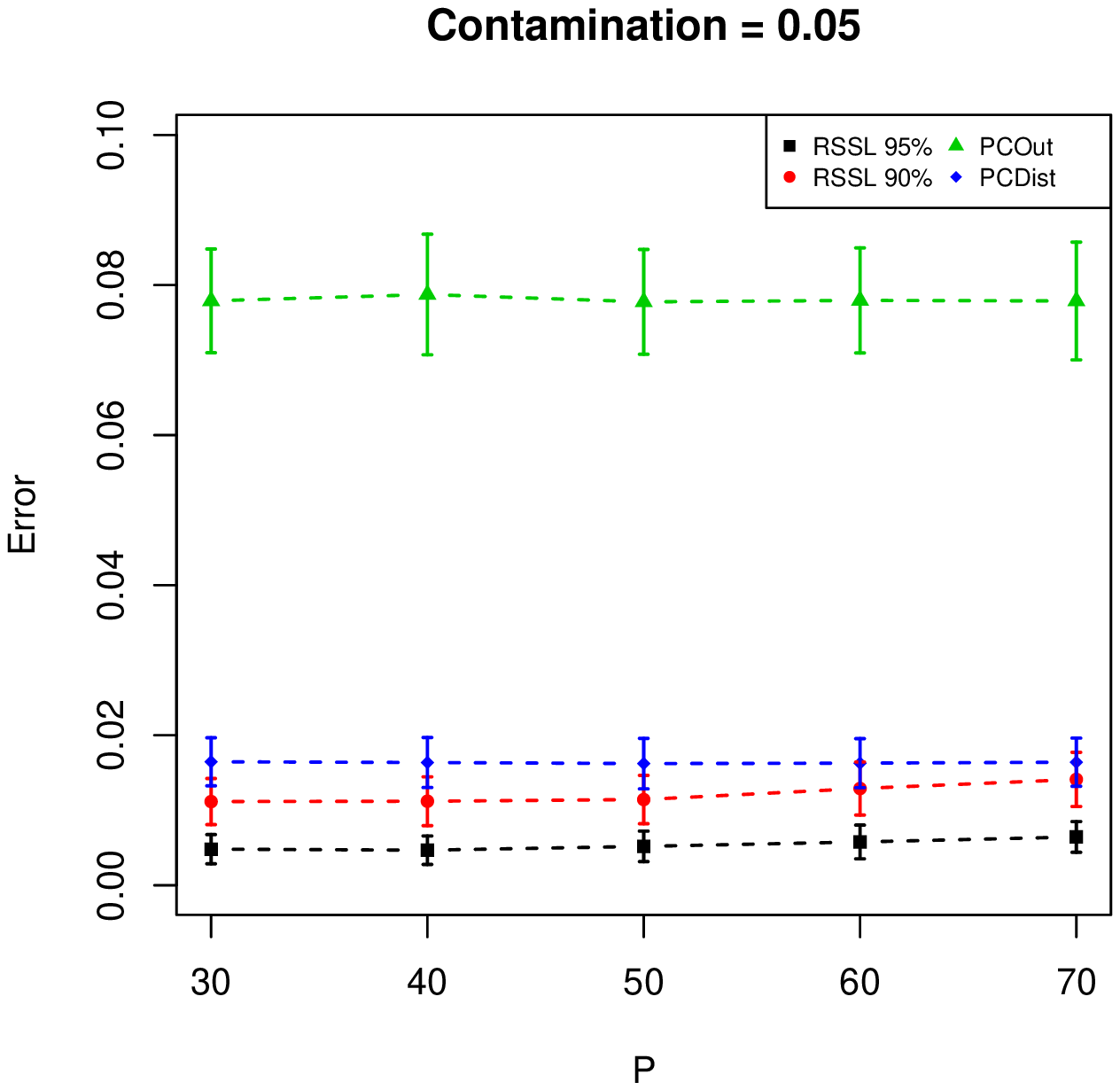, width=7cm,height=7cm}
  \epsfig{figure=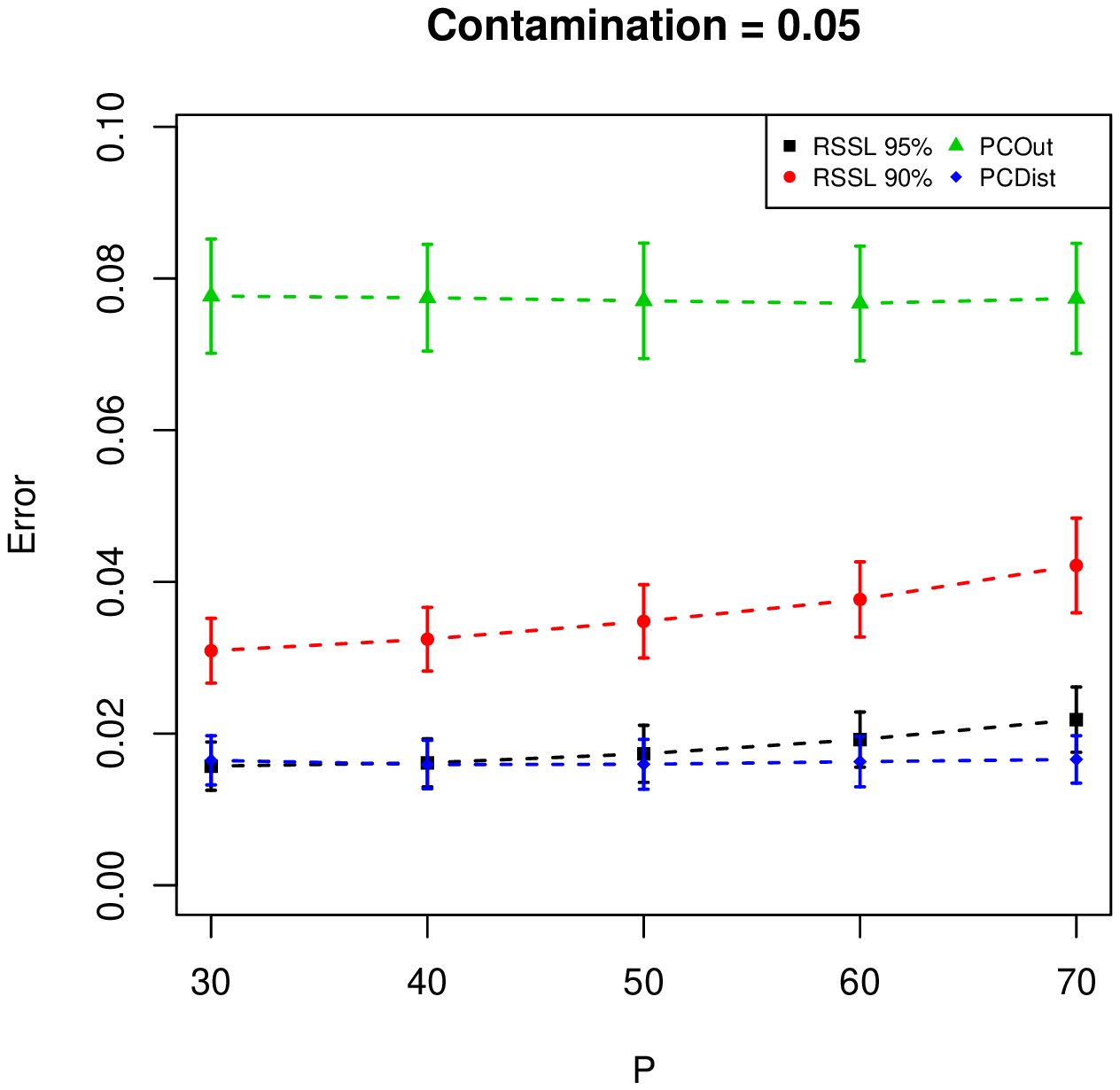, width=7cm,height=7cm}
  \epsfig{figure=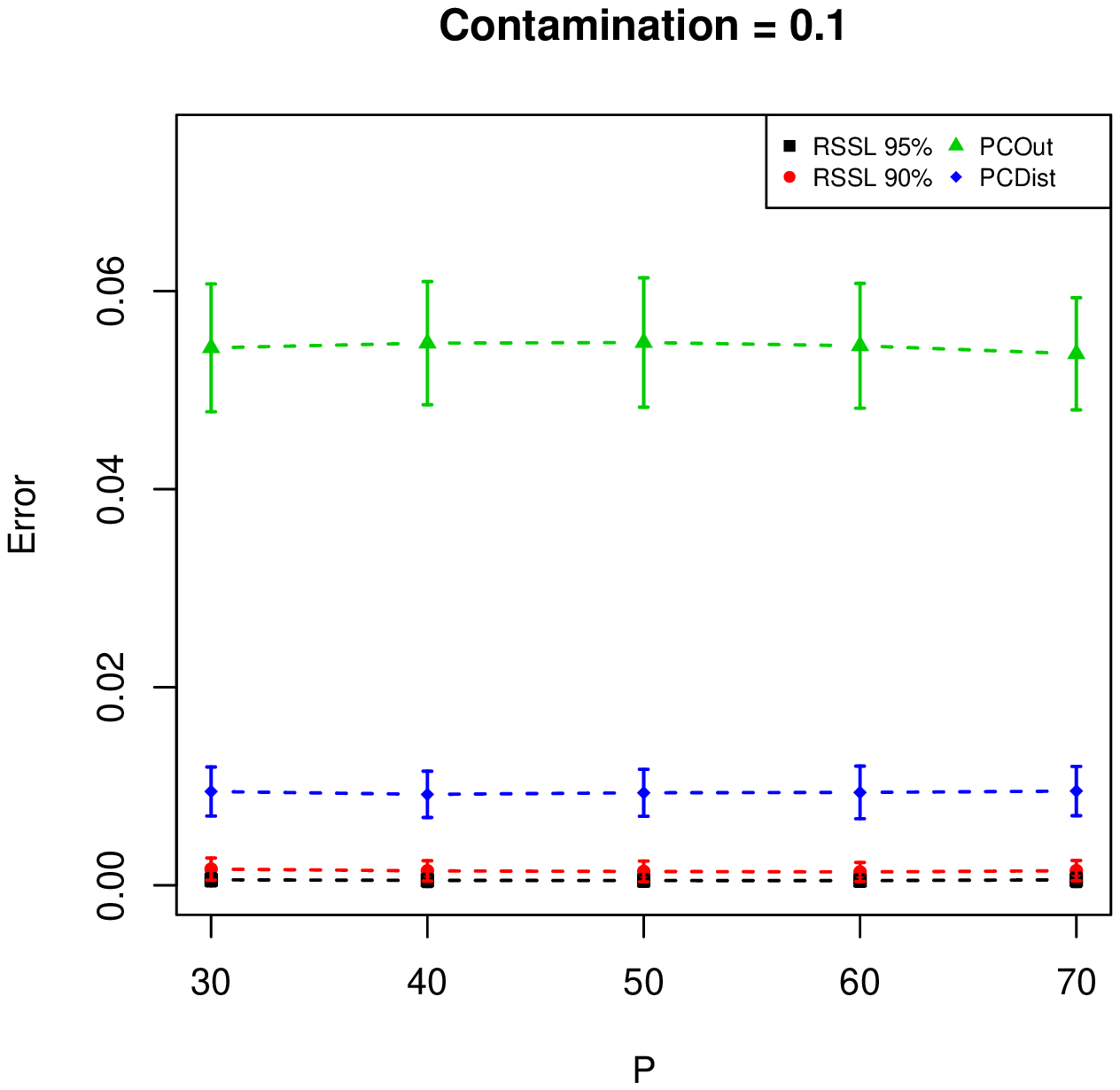, width=7cm,height=7cm}
  \epsfig{figure=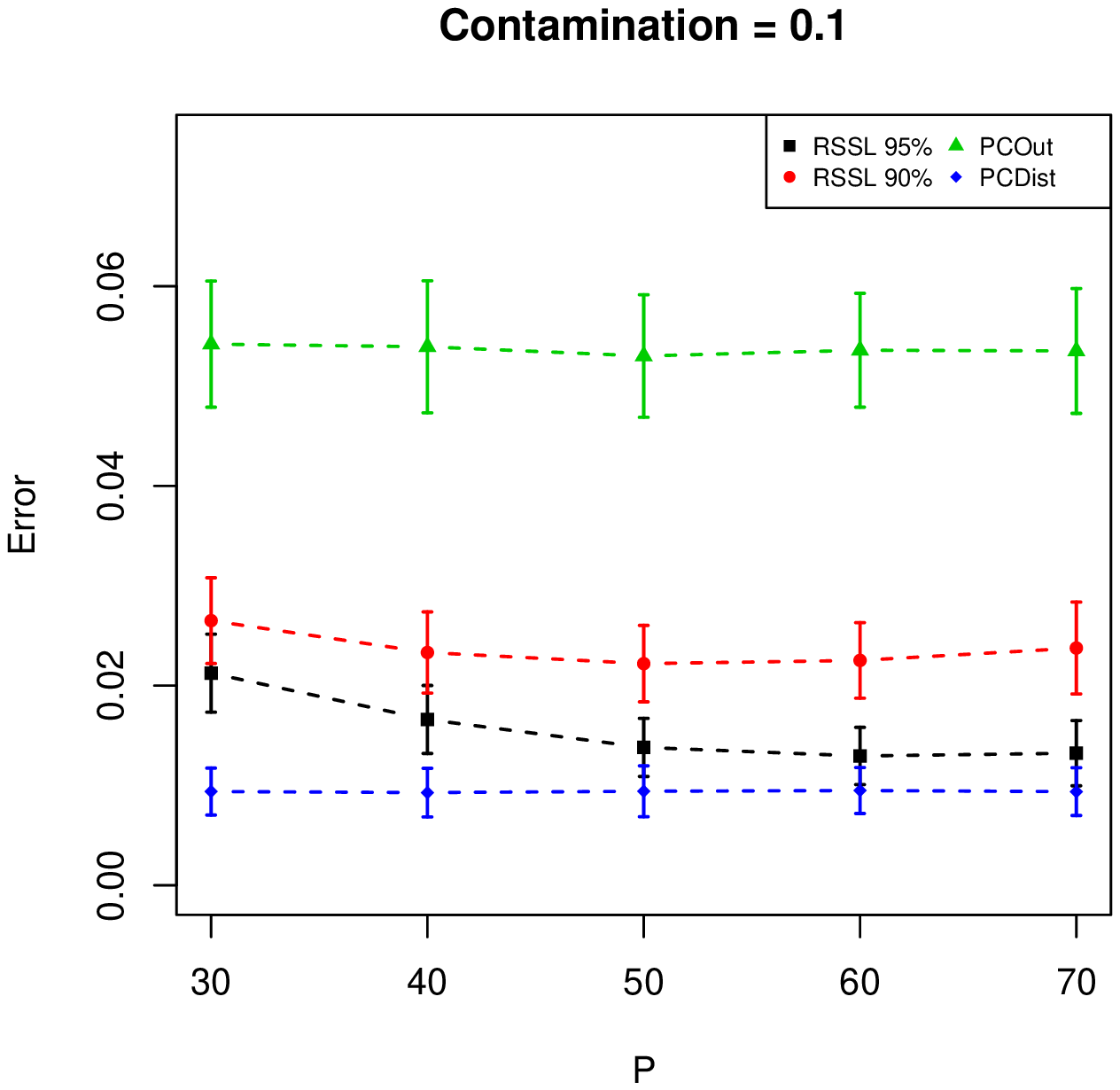, width=7cm,height=7cm}
  \epsfig{figure=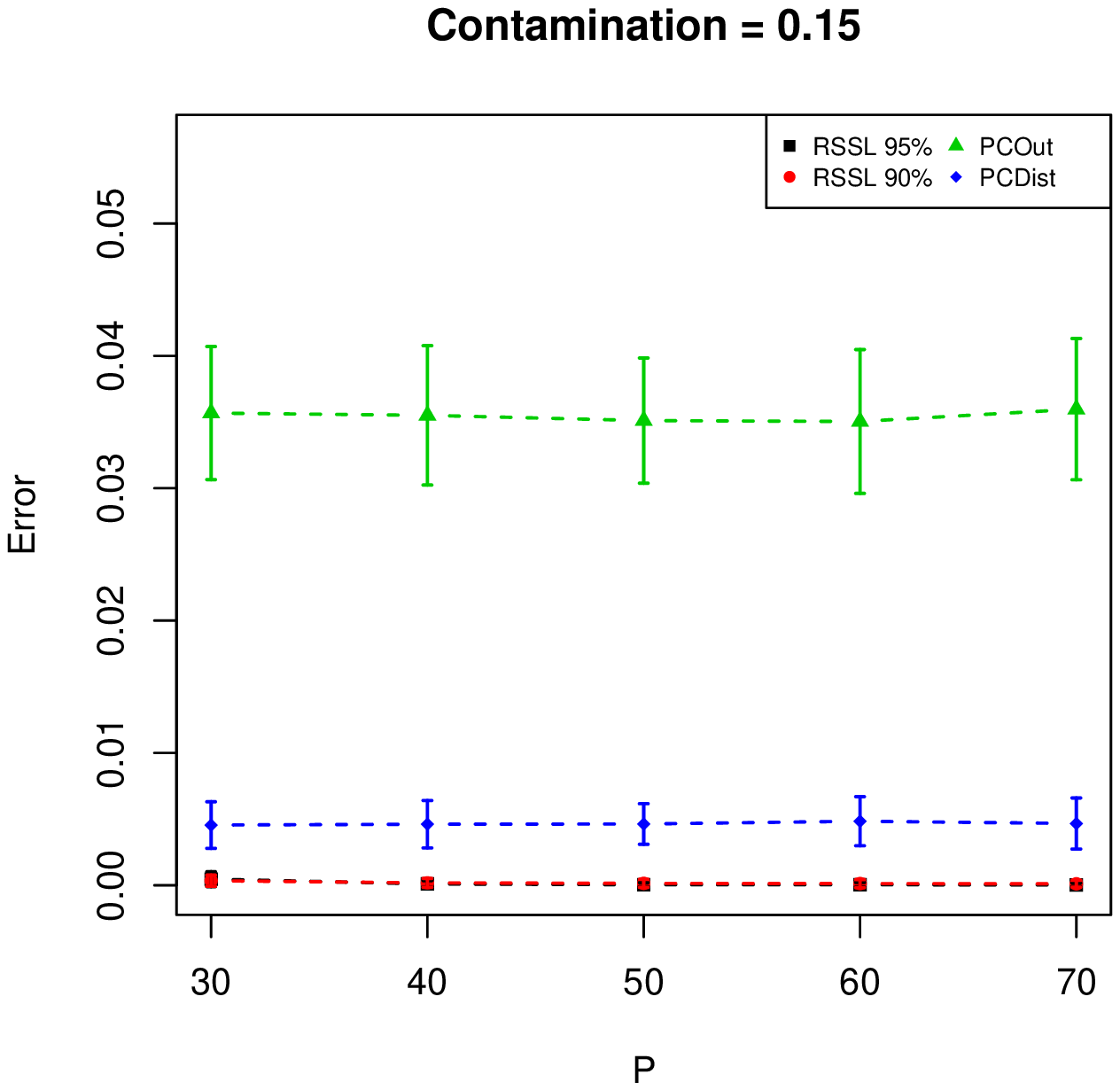, width=7cm,height=7cm}
  \epsfig{figure=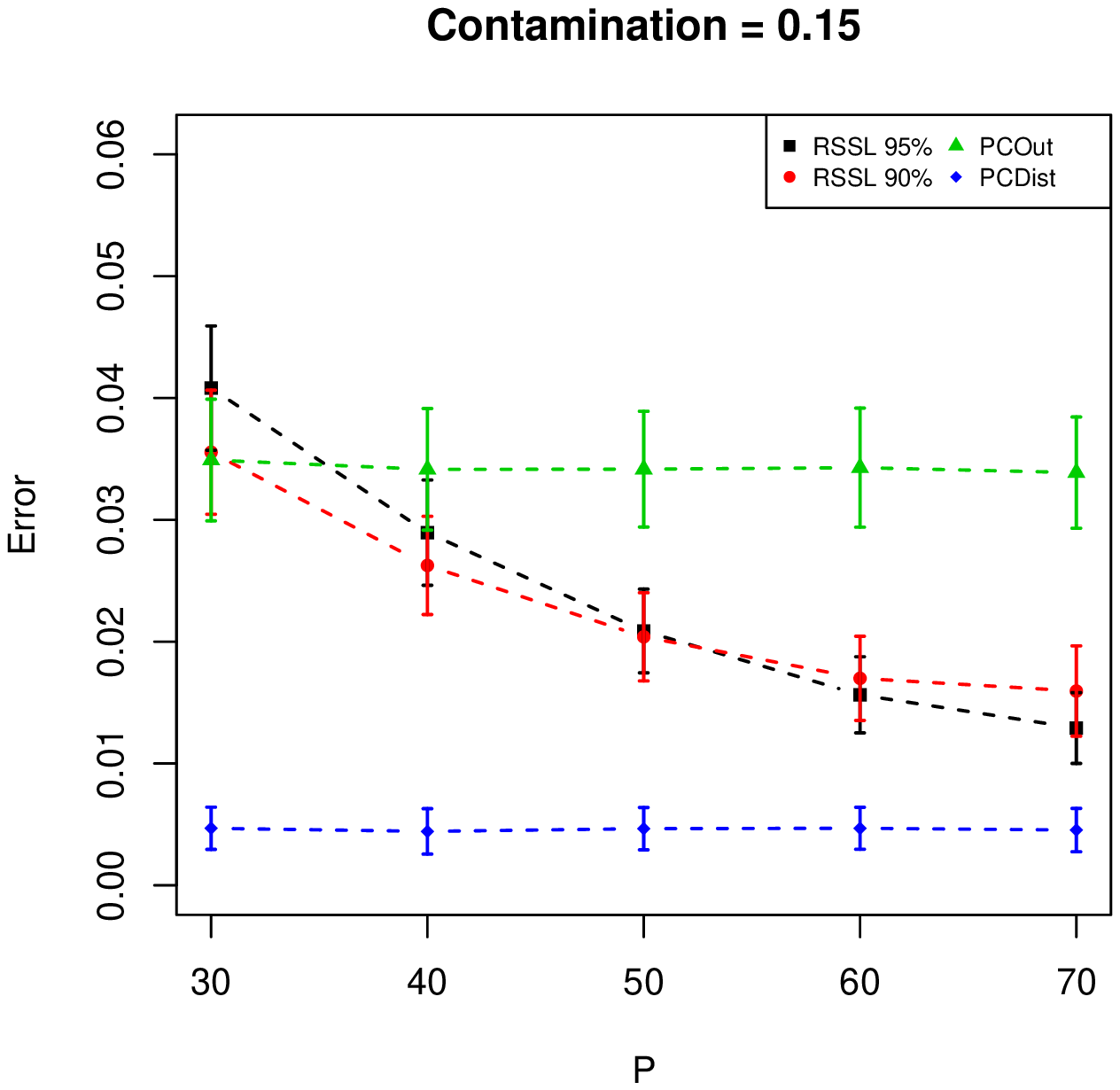, width=7cm,height=7cm}
  \caption{The average error and standard deviation in low dimensional simulation with $\kappa, \eta = 5$ (left column) and $\kappa, \eta = 2$ (right column).}
  \label{fig:4}
\end{figure}

\begin{figure}[H]
\centering
  \epsfig{figure=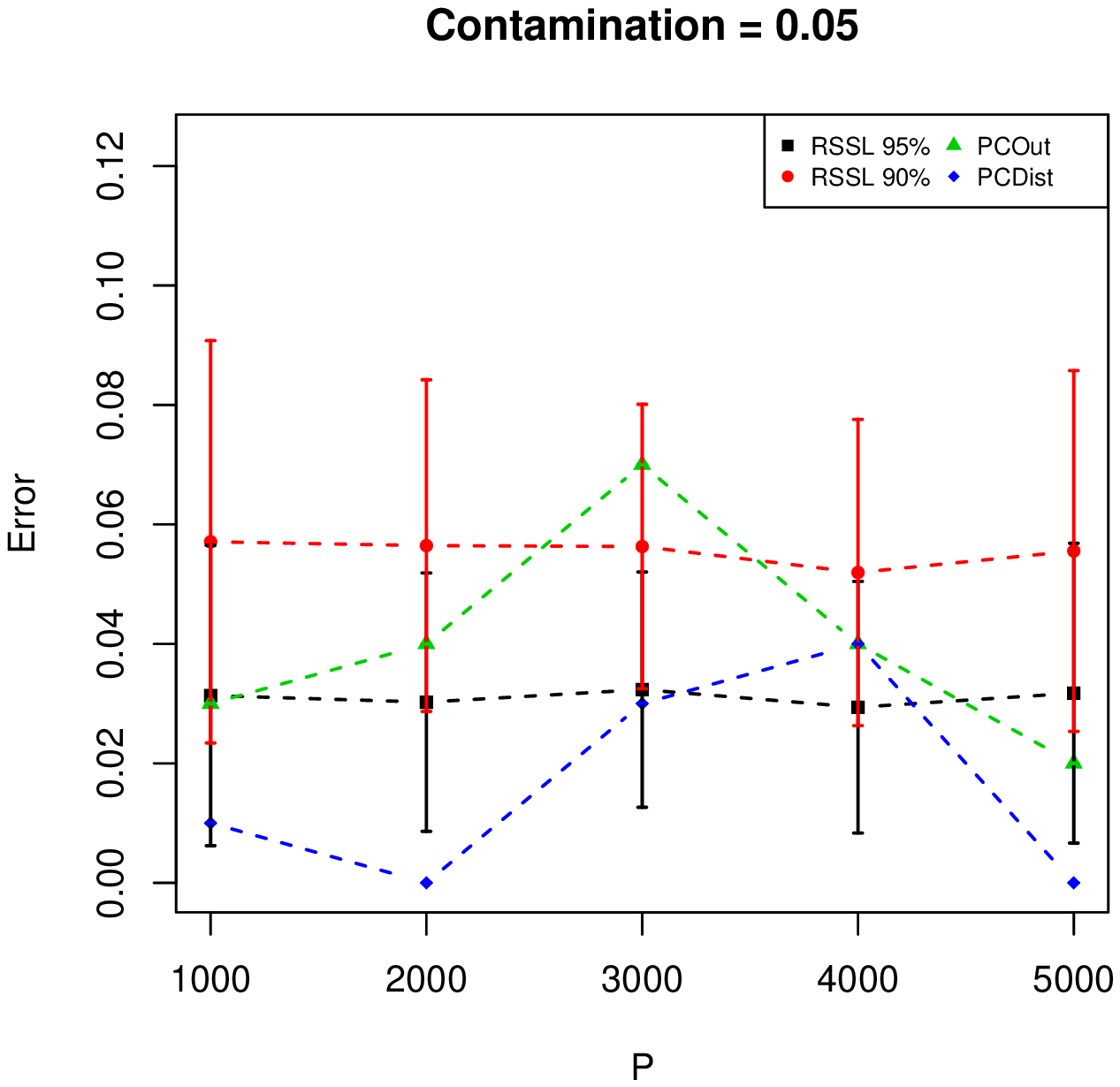, width=7cm,height=7cm}
  \epsfig{figure=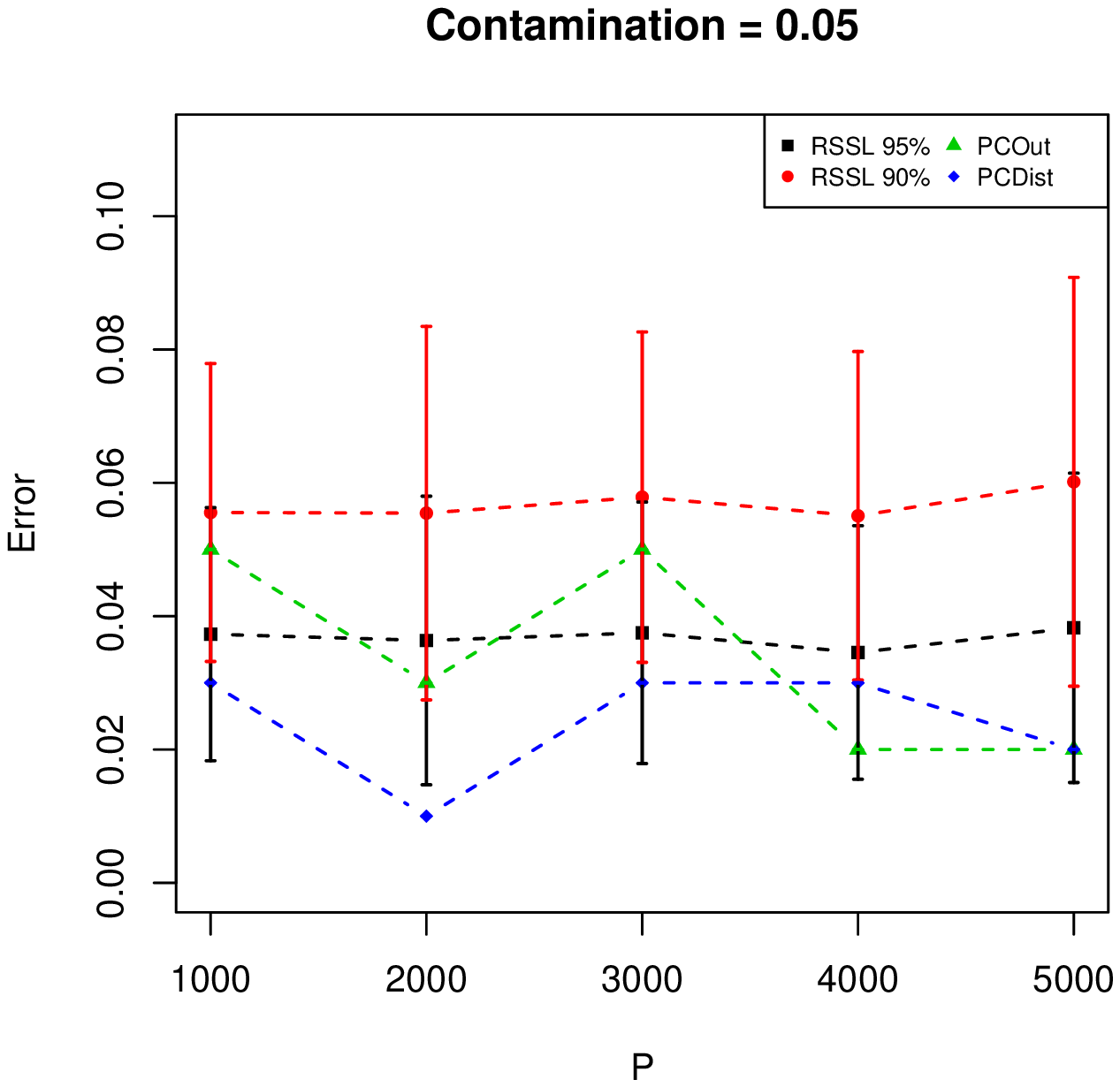, width=7cm,height=7cm}
  \epsfig{figure=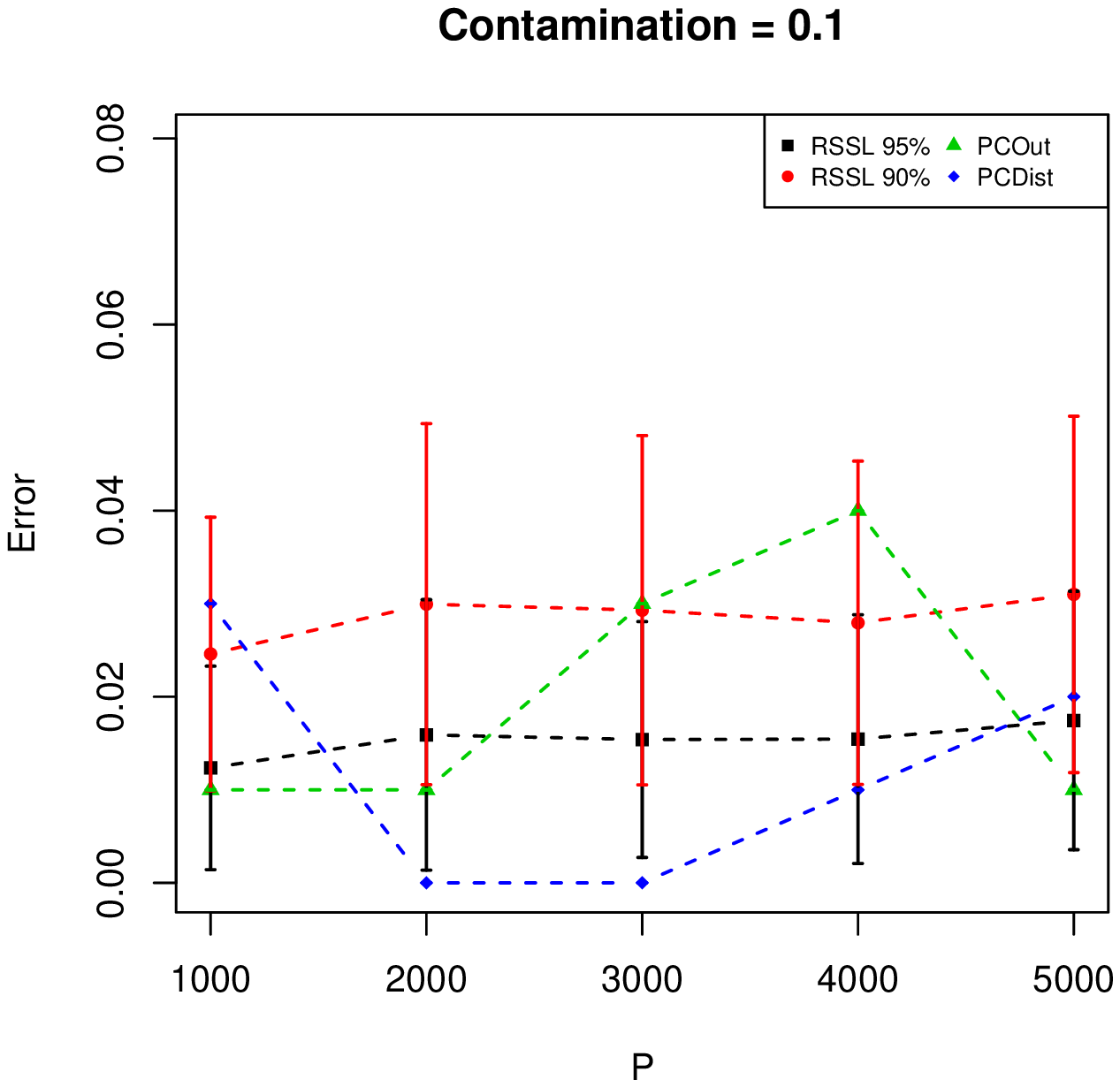, width=7cm,height=7cm}
  \epsfig{figure=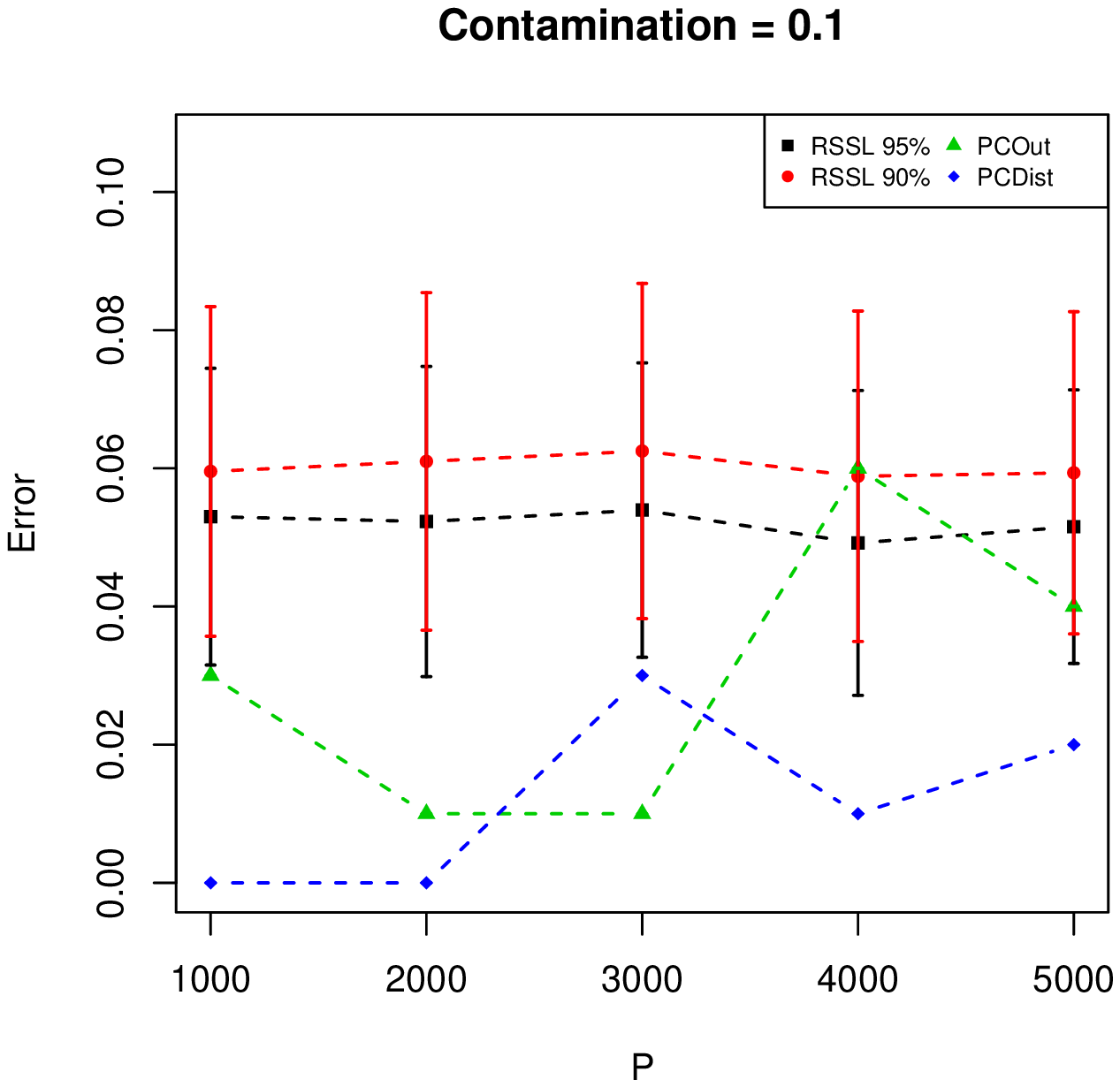, width=7cm,height=7cm}
  \epsfig{figure=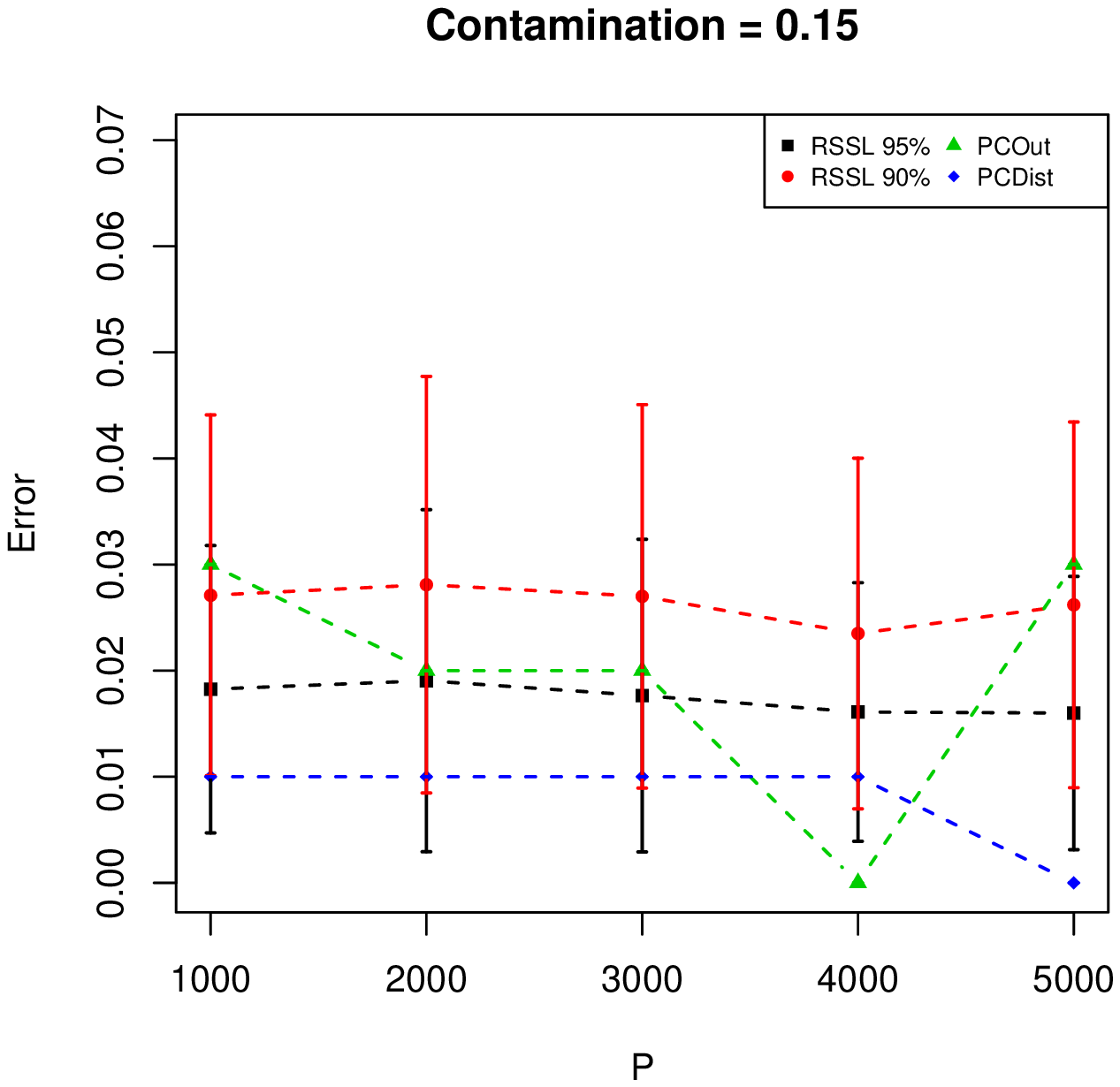, width=7cm,height=7cm}
  \epsfig{figure=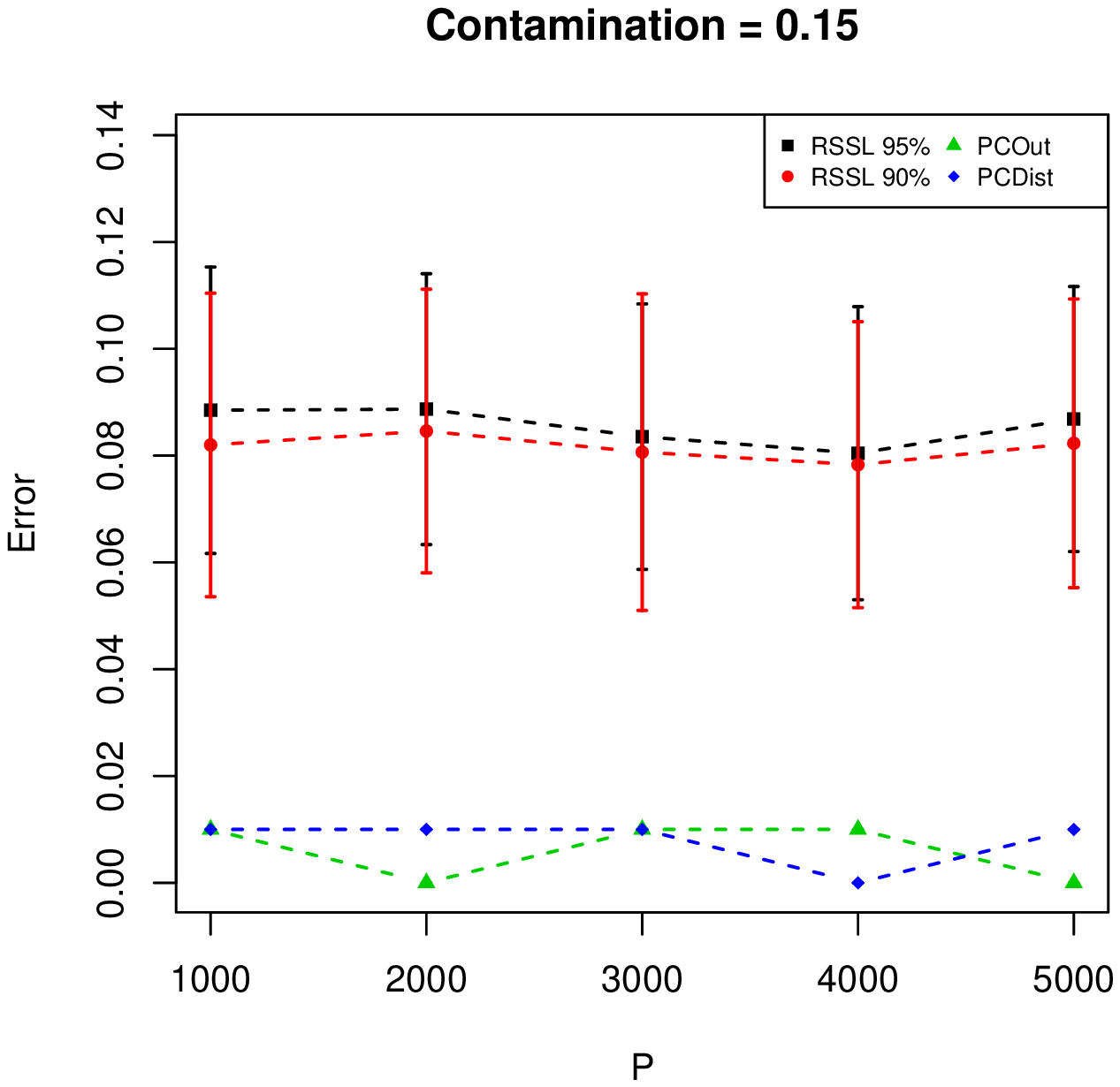, width=7cm,height=7cm}
  \caption{The average error and standard deviation in high dimensional simulation with $\kappa, \eta = 5$ (left column) and $\kappa, \eta = 2$ (right column).}
  \label{fig:5}
\end{figure}

The improvement of our random subspace learning algorithm in low dimensional data with $p \in \{30, 40, 50, 60, 70\}$ and relative large sample size $n = 1500$, is demonstrated in figure \eqref{fig:4} in comparison to $\tt{PCOut}$ and $\tt{PCDist}$. Given a relatively easy task, namely with $\kappa, \eta = 5$, the outliers are scattered widely and shifted far from normal, the RSSL with $1-\alpha$ equals $95\%$ and $90\%$ perform consistently very well, typically outperforming the competition. When the rate of contamination is increasing in this scenario, almost $100\%$ accuracy can be achieved with RSSL based algorithm. When the outliers are spread more narrowly and closer to the mean with $\kappa, \eta = 2$, the predictive accuracy of our random subspace based algorithm is slightly less powerful but still very strong, namely with a predictive detection rate close to $96\%$ to $99\%$. In high dimensional settings,  namely with $p \in \{1000, 2000, 3000, 4000, 5000\}$ and low sample size $n = 100$, RSSL is also performs reasonably well as shown in figure \eqref{fig:5}. With $1-\alpha = 95\%$ chi-squared cut-off, when $\kappa, \eta = 5$, $96\%$ to $98\%$ of outliers can be detected constantly among all simulated high dimensions. Under more difficult conditions, as with $\kappa, \eta = 2$, a decent amount of outliers can be detected with accuracy around $92\%$ to $96\%$. Based on the properties of robust PCA based algorithms, the situation that we define as "easy" for RSSL algorithms is actually "harder" for $\tt{PCOut}$ and $\tt{PCDist}$. The principle component space is selected based on the visibility of outliers, and especially for $\tt{PCOut}$, the components with nonzero robust kurtosis are assigned higher weights by the absolute value of their kurtosis coefficients. This method is shown to yield good performances when dealing with small shift of mean and scatter of the covariance matrix. However, if the outliers lied on larger $\eta$ and $\kappa$ where excessive choices can be made then, it is more difficult for PCA to find the dimensionality to make the outliers "stick out". Reversely, with a small values of $\kappa$ and $\eta$, the most obvious directions are emphasized by PCA but less chance for algorithms like RSSL to obtain the most sensible subspace to build robust estimators. So in figure \eqref{fig:5}, when $\kappa, \eta = 2$ the accuracy reduced to around $92\%$ but in all other high-dimensional settings the performance of RSSL is consistent with $\tt{PCOut}$ and identically stable.

\section{Conclusion}

We have presented what we can rightfully claim to be a computational efficient, scalable, intuitive appealing and highly predictively accurate outlier detection method for both HDLSS and LDHSS datasets. As an adaptation of both random subspace learning and
 minimum covariance determinant, our proposed approach can be readily used on vast number of real life examples where both its component
 building blocks have been successfully applied. The particular appeal of the random subspace learning aspect of our method comes in handy for many outlier detection tasks on high dimension low sample size datasets like DNA Microarray Gene Expression datasets for which the MCD approach proved to be computational untenable. As our computational demonstrations section above reveal, our proposed approach competes favorably with other existing methods, sometimes outperforming them predictively despite its straightforwardness and relatively simple implementation.
 Specifically, our proposed method is shown to be very competitive for both low dimensional space and high dimensional space outlier detection and is computationally very efficient. We are currently seeking out interesting real life datasets on which to apply our method. We also plan to extend our method beyond settings where the underlying distribution is Gaussian.

\section*{Acknowledgements}
Ernest Fokou\'e wishes to express his heartfelt gratitude and infinite thanks to Our Lady of Perpetual Help for Her
ever-present support and guidance, especially for the uninterrupted flow of inspiration received through Her
most powerful intercession.

\section*{References}

\bibliography{lf-robust-ref}
\end{document}

%% file: Macros.tex
\newcommand{\nmathbf}{\bm}

\def\bfI{\nmathbf I}

\def\bfV{\nmathbf V}

\def\bfmu     {\nmathbf \mu}

\def\bfxi     {\nmathbf \xi}

\def\bfLambda {\nmathbf \Lambda}

\def\bfSigma  {\nmathbf \Sigma}

\def\boldfacefake#1{\kern-4pt
    \hbox{ \mathsurround=0pt
    \hbox to 0.4pt{$#1$\hss}\hbox to 0.4pt{$#1$\hss}\hbox {$#1$}}}

\newcommand{\be}{\begin{eqnarray}}
\newcommand{\ee}{\end{eqnarray}}
\newcommand{\ba}{\begin{eqnarray*}}
\newcommand{\ea}{\end{eqnarray*}}

\newcommand{\reals}{\mbox{\rm I\kern-.20em R}}
\newcommand{\sreals}{\mbox{\small \rm I\kern-.20em R}}